\documentclass{article} 
\usepackage{iclr2022_conference,times}

\usepackage{amsmath,amsfonts,bm}

\def\eqref#1{equation~\ref{#1}}

\def\1{\bm{1}}

\DeclareMathAlphabet{\mathsfit}{\encodingdefault}{\sfdefault}{m}{sl}
\SetMathAlphabet{\mathsfit}{bold}{\encodingdefault}{\sfdefault}{bx}{n}

\newcommand{\E}{\mathbb{E}}

\newcommand{\R}{\mathbb{R}}

\input{include}

\title{Generalized rectifier wavelet covariance models for texture synthesis}

\iclrfinalcopy 
 
\author{Antoine Brochard \\
ENS, PSL University, Paris, France\\
\texttt{antoine.brochard@ens.fr} \\
\And
Sixin Zhang\\
Université de Toulouse, INP, IRIT, Toulouse, France  \\
\texttt{sixin.zhang@irit.fr} \\
\And
St\'ephane Mallat \\
Coll\`ege de France, Paris, France \\
Flatiron Institute, New York, USA\\
}

\begin{document}

\maketitle

\begin{abstract}

State-of-the-art maximum entropy models for texture synthesis
are built from statistics relying on image representations defined 
by convolutional neural networks (CNN). 
Such representations capture rich structures in texture images, 
outperforming wavelet-based representations in this regard. 
However, conversely to neural networks, 
wavelets offer meaningful representations, 
as they are known to detect structures at multiple scales (e.g. edges) in images.
In this work, we propose a family of statistics built upon 
non-linear wavelet based representations, 
that can be viewed as a particular instance of a one-layer CNN, using
a generalized rectifier non-linearity.
These statistics significantly improve the visual quality of previous classical wavelet-based models, and allow one to produce syntheses of similar quality 
to state-of-the-art models, on both gray-scale and color textures.
We further provide insights on memorization effects in these models. 
\end{abstract}

\section{Introduction}\label{s.intro}

Textures ares spatially homogeneous images,
 consisting of similar patterns forming a coherent ensemble.
In texture modeling, 
one of the standard approaches to synthesize textures
relies on defining a maximum entropy model \citep{jaynes1957information}
using a single observed image \citep{raad2018survey}. 
It consists of computing a set of prescribed statistics from the observed texture image,
and then generating synthetic textures producing the same statistics as the observation. 
If the statistics correctly describe the structures present in the observation, 
then any new image with the same statistics should appear similar to the observation.
A major challenge of such methods resides in finding a suitable set of statistics, that 
can generate both high-quality and diverse synthetic samples. 
This problem is fundamental as it is at the heart of many texture related problems.
For example, 
in patch re-arrangement methods for texture modeling, these statistics
are used to compute high-level similarities of image patches \citep{li2016combining,raad2018survey}.
Such models are also used
for visual perception
\citep{freeman2011metamers,wallis2019image,vacher2020texture}, 
style transfer \citep{gatys2016image,deza2018towards} 
and image inpainting \citep{laube2018image}.

A key question along this line of research is to find
\textit{what it takes to generate natural textures.}
This problem was originally posed in \citet{julesz1962visual}, in which the author
looks for a statistical characterization of textures.
In the classical work of \citet{portilla2000parametric} (noted PS in this work), 
the authors presented a model whose 
statistics are built on the wavelet transform of an input texture image. 
These statistics were carefully chosen, by showing that each of them captured a specific 
aspect of the structure of the image. 
This model produces satisfying results for a wide range of textures,
but fails to reproduce complex geometric structures 
present in some natural texture images.  
\Cref{fig:ps} presents a typical example composed of radishes, 
and synthetic images from three state-of-the-art models
developed over the last few decades. 
To address this problem, 
the work of \citet{gatys2015texture} proposes to use statistics built on the correlations 
between the feature maps of a 
deep CNN, pre-trained on the ImageNet
classification problem \citep{deng2009imagenet, simonyan2014very}. 
While this model produces visually appealing images, these statistics are hard to interpret. 
The work of \citet{ustyuzhaninov2017what} made a 
significant simplification of 
such statistics, 
by using the feature maps of a one-layer rectifier CNN 
with random filters (without learning). 
A crucial aspect of this simplification relies on using multi-scale filters, 
which are naturally connected to the wavelet transform. 
In this paper, 
we propose a wavelet-based model, more interpretable than CNN-based models (with learned or random filters), 
to synthesize textures with complex geometric structures. It allows to bridge the gap between the classical work of \citet{portilla2000parametric}, and state-of-the-art models. 

{\setlength{\tabcolsep}{2pt}
\begin{figure}[!h]
    \centering
    \begin{tabular}{cccc}
    Observation & PS & VGG & RF   \\
    \includegraphics[width=0.17\linewidth]{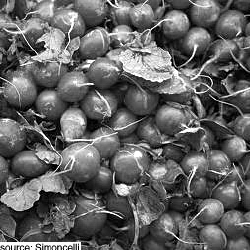}&
    \includegraphics[width=0.17\linewidth]{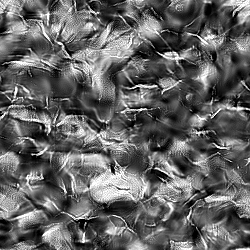}&
    \includegraphics[width=0.17\linewidth]{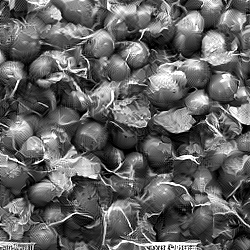}&
    \includegraphics[width=0.17\linewidth]{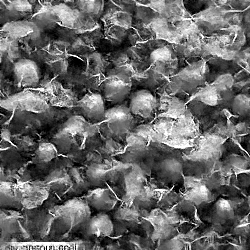}
    \end{tabular}
    \caption{Example of syntheses from three texture models in chronological order. From left to right: the observed texture in gray-scale, synthesis from PS \citep{portilla2000parametric}, from VGG \citep{gatys2015texture}, and from RF\protect\footnotemark \citep{ustyuzhaninov2017what}.}
    \label{fig:ps}
\end{figure}}

\footnotetext{As the model from \citet{ustyuzhaninov2017what} uses on random filters, we shall use the abbreviation RF.}

This model is
built on the recent development of the phase harmonics 
for image representations 
and non-Gaussian stationary process modeling \citep{mallat2020phase,zhang2021maximum}. The phase harmonics are non-linear transformations 
that adjust the phase of a complex number. 
In \citet{portilla2000parametric,zhang2021maximum}, 
the authors illustrate 
that the phase dependencies between wavelet coefficients 
across scales contain important information about the geometric structures in textures and turbulent flows, 
and that they can be captured by applying the phase harmonics to
complex wavelet coefficients. 
Remarkably, \citet{mallat2020phase} show that the phase harmonics
admit a dual representation, closely related 
to the rectifier non-linearity in CNNs.
Our main contributions are: 

\begin{itemize}
    \item We develop a family of texture models based on the wavelet transform and a generalized rectifier non-linearity, that significantly improves the visual quality 
    of the classical wavelet-based model of \citet{portilla2000parametric}
    on a wide range of textures. 
    It relies on introducing spatial shift statistics across scales 
    to capture geometric structures in textures. 
    \item  By changing the number of statistics in our models, 
    we show explicitly the trade-off on the quality and diversity of the synthesis.
    When there are too many statistics, our model tends to memorize image patches.
    We further investigate such memorization effects on non-stationary images and find that 
    it sometimes relies on what statistics are chosen, rather than on how many.
    \item Through the modeling of geometric structures in gray-scale textures, 
    our model indicates the possibility of reducing 
    significantly the number of statistics
    in the works of \citet{gatys2015texture} and \citet{ustyuzhaninov2017what}, 
    to achieve a similar visual quality.
\end{itemize}

The rest of the paper is organized as follows: 
Section \ref{model:cov} reviews the framework of microcanonical maximum-entropy models, 
build upon a general family of covariance statistics. 
We then present our model for both gray-scale and color textures in Section \ref{s.shift}. Section \ref{s.results} shows synthesis results of our model, compared with state-of-the-art models. Finally, in \Cref{s.discussion}, we discuss possible improvements of our model\footnote{ All calculations can be reproduced by a Python software
available at \url{https://github.com/abrochar/wavelet-texture-synthesis}.}.

\paragraph{Notations}
Throughout the paper, 
$N$ denotes a positive integer. A gray-scale image $x$ is an element of $\R^{N \times N}$, i.e. $x=x(u), \: u \in \Omega_N$, with $\Omega_N := \{0, \cdots, N-1\}^2$. 
A color image $x = \{x^c\}_{c=1,2,3}$ is an element of 
$\R^{3 \times N \times N}$, or equivalently, each $ x^c \in \R^{N \times N}$. 
We shall denote $\bar{x}$ the observed texture (observation),
which is assumed to be a realisation a random vector $X$. 
For any complex number $z \in \mathbb{C}$, $z^\ast$ is the complex conjugate of $z$, $\mbox{Real}(z)$ its real part, $|z|$ its modulus, and $\varphi(z)$ its phase.

\section{Microcanonical covariance models}
\label{model:cov}

We briefly review the standard framework 
of micro-canonical maximum-entropy models 
for textures. To reliably 
estimate the statistics in these models, 
we assume that a texture is a realization
of a stationary and ergodic process $X$
(restricted to $\omN$). 
We then review a special family of statistics
that are used in the state-of-the-art texture models (mentioned in \Cref{fig:ps}), 
based on covariance statistics of an image representation.

\subsection{Framework}\label{ss.framework}

Given a observation texture $\bar{x}$, we aim at generating new texture images, similar but different from $\bar{x}$.
To that end, a classical method is to define a set of statistics $C\bar{x}$, computed on the observation, 
and try to sample from the {\em microcanonical set}
\begin{equation*}
    \{ x : \| Cx - C\bar{x} \| \leq \epsilon \},
\end{equation*}
where $\|\cdot\|$ denotes the $L^2$ norm. 

Under the stationary and ergodic assumption of $X$, 
one can construct $ C x$ as a statistical estimator of $\E(CX)$, 
from a complex-valued representation $Rx$.\footnote{The complex-valued representation $ R x (\ga,u)  \in \C $ 
is a function of $ (\ga,u)  $ in an index set $\Ga \times \omN$.}
The set of covariance statistics $Cx$ of a model can then be constructed by 
computing an averaging over the spatial variable $u$, i.e. 
\begin{equation}\label{eq:corr}
    C x(\gamma, \ga', \tau) := \frac{1}{|\omN|} \sum_{u \in \omN} Rx (\ga, u) Rx (\ga',u-\tau)^\ast ,
\end{equation}
for $(\ga, \ga', \tau) \in \U \subseteq \Ga \times \Ga \times \omN$.
The statistics $C x(\gamma, \ga', \tau) $ 
can be interpreted as 
estimating the covariance (resp. correlations)
between $RX(\ga,u)$ and $RX(\ga',u-\tau)$
for zero-mean $RX$ (resp. non-zero mean $RX$).
The ergodicity assumption ensures that when $N$ is large enough, 
the approximation $ C\bar{x}  \simeq \E ( C X )$ over $\U$ should hold with high probability. Under these conditions, it makes sense to sample the microcanonical set in order to generate new texture samples.


This framework encompasses a wide range of state-of-the-art texture models\footnote{e.g. \citet{portilla2000parametric, gatys2015texture, ustyuzhaninov2017what, zhang2021maximum}}. 
In particular, the PS model takes inspiration from the human early visual system to define a multi-scale representation based on the {\em wavelet transform} of the image \citep{heeger1995pyramid}. 
We next review a family of covariance model which 
generalizes the statistics in the PS model. We write $C^\m$ the statistics 
for a specific model $\m$ that uses the representation $R^\m$.


\subsection{Wavelet phase harmonic covariance models}

We review a family of microcanonical 
covariance models defined by a representation 
built upon the wavelet transform and phase harmonics. 
It defines a class of covariance statistics 
that capture dependencies between wavelet coefficients across scales. 


\subsubsection{Wavelet transform}\label{ss.wavelet-transform}

The wavelet transform is a powerful tool in image processing to 
analyze signal structures, by defining a sparse representation \citep{mallatbook}.
For texture modeling, we consider 
oriented wavelets to model 
geometric structures in images at multiple scales. 
They include the Morlet wavelets and steerable wavelets, 
proposed in \citet{GOUPILLAUD198485,steerableSimoncelli,unseral_steerable13}.
In particular, the Simoncelli steerable wavelets 
have been used to model a diverse variety of textures in \citet{portilla2000parametric}. 

Oriented wavelets 
are defined by the dilation and rotation 
of a complex function $\psi:\R^2 \mapsto \mathbb \C$ on a plane.
Let $r_{\theta}$ denote the rotation by angle $\theta$ in $\Rdd$. 
They are derived from $\psi$ with dilations
by factors $2^j$, for $j \in \{0,1,\cdots,J-1\}$, 
and rotations $r_\theta$ over
angles $\theta = \ell \pi / L$ for $0 \leq \ell < L$, 
where $L$ is the number of angles in $[0, \pi)$. 
The wavelet at scale $j$ and angle $\theta$ is defined 
by 
\[
    \psi_{j,\theta} (u) =      2^{-2 j} \psi (2^{-j} r_{  \theta  } u ) , \quad  u \in \R^2. 
\]
Scales equal or larger than $J$ are carried by a low-pass filter $\phi_J$. 

The wavelet transform of an image 
$x \in \R^{N \times N}$ is a family of functions 
obtained by the convolution of $x$ with 
discrete wavelets.\footnote{
The continuous wavelets are discretized with periodic boundary conditions on the spatial grid $\omN$. 
}
Let $\La := \{0, \cdots, J-1\} \times \frac{\pi}{L}\{0, \cdots, L-1\}$ be an index set.
The wavelet coefficients are
\begin{equation}\label{periodic}
    x \star \psi_{j,\theta}  (u) = 
    \sum_{v \in \omN} x(u-v) \psi_{j,\theta} (v) , \quad u \in \omN,  \quad (j,\theta) \in \La  . 
\end{equation}
The low-pass coefficients $x \star \phi_J $ are defined similarly.

\subsubsection{Wavelet phase harmonics and the PS model}

To model natural textures, 
it has been shown \citep{portilla2000parametric, zhang2021maximum} that
it is crucial to capture statistical dependencies between wavelet coefficients
across scales. 
This can be achieved by using a 
wavelet phase harmonic representation, 
which is defined by the composition of a linear wavelet transform of $x$, 
and a non-linear phase harmonic transform. 

In \citet{mallat2020phase}, the authors introduce
the phase harmonics to adjust 
the phase of a complex number $z \in \C$. 
More precisely, the phase harmonics $\{ [z]^k \}_{k \in \Z}$ 
of a complex number $z \in  \mathbb C$ are defined 
by multiplying its phase $\varphi(z)$ of $z$ 
by integers $k$, while keeping the modulus constant, i.e. 
$\forall \: k \in \mathbb Z, \: [z]^k := |z|e^{ik\varphi(z)}.$ 
The wavelet phase harmonic representation (WPH) is then defined  by
\begin{equation}
    R^{\WPH} x  (\ga,u) = [ x \star \psi_{j,\theta} (u) ]^k - \mu_{\ga},
    \quad 
    \ga = (j, \theta, k) \in \Ga = \La \times \Z, 
\end{equation}
where $\mu_\ga$ is defined as the spatial average of $ [ \bar{x} \star \psi_{j,\theta} ]^k$.

It is shown in \citet{zhang2021maximum} that the PS model 
can be regarded as a low-order wavelet phase harmonics covariance model, which considers only a restricted number of pairs $(k, k')$ (see \Cref{app:params} for more details).
In the next section, we shall use a dual representation 
of the phase harmonic operator to define a covariance model to capture high-order phase harmonics. 





\section{Generalized rectifier wavelet covariance model}\label{s.shift}

In the previous section, we presented a class of models, built from the wavelet phase harmonic representation. 
A dual representation of the phase harmonic operator $[\cdot]^k$ 
can be defined via a generalized rectified linear unit, that we review in Section \ref{ss.k-to-alpha}. 
We then discuss in Section \ref{ss.extremes} 
how to define an appropriate index set of $\Gamma$ for gray-scale textures. 
Section \ref{sec:color} extends the model to color textures.

\subsection{From phase harmonics to the generalized rectifier}\label{ss.k-to-alpha}

The generalized rectified linear unit of a complex number $z$, with a phase shifted by $\al \in [0, 2\pi]$, is defined by
\begin{equation}\label{alpha-relu}
    \rho_\al (z) = \rho ( \mbox{Real} ( e^{i \alpha } z ) ),
\end{equation}

where $\rho$ is a rectified linear unit, i.e. for any $t \in \R$, $\rho(t) := \max(0,t)$.
In \citet{mallat2020phase}, it is shown that applying a Fourier transform on $\rho_\al(z)$ along the variable $\alpha $ results in the phase harmonics of $z$ (up to some normalization constant). This suggests an alternative model, defined by coefficients of the form 
\begin{equation}\label{alpha-features}
    R^{\ALPHA} x  (\ga, u) = \rho_\alpha (  x \star \psi_{j,\theta} (u) ) - \mu_{\ga},  \quad \ga = (j, \theta, \al) , 
\end{equation}
for $\ga \in \Ga = \La \times [0, 2\pi]$, and $\mu_\gamma$ is defined as the spatial average of  $  \rho_\alpha (  \bar{x} \star \psi_{j,\theta} (u) )$  over $u \in \omN$.

\paragraph{Relation with high-order phase harmonics}
Based on the duality between the phase harmonics $k \in \Z$ 
and the phase shift variable $\alpha \in [0, 2\pi]$, 
we now present the relation between 
$C^\ALPHA$ and the high-order phase harmonics 
in $C^\WPH$, first proved in \citet{mallat2020phase}.
\begin{prop}\label{prop.equiv}
    There exists a complex-valued sequence $\{ c_k \}_{k \in \Z}$ such that for all $(j,\theta,\alpha) \in \Ga$, $(j',\theta',\alpha') \in \Ga$, and all $\tau \in \omN$, 
    \[
        C^\ALPHA x (( j,\theta,\alpha), (j',\theta',\alpha'),\tau ) 
        = \sum_{(k,k') \in \Z^2}  c_k c_{k'}^\ast 
            C^\WPH x  (( j,\theta, k ), (j',\theta',k'),\tau )  e^{i (k \alpha - k'\alpha')}  . 
    \]
\end{prop}

The proof is given in \Cref{proof-equiv}. 
We remark that the sequence $\{ c_k \}_{k \in \Z}$ is uniquely 
determined by the rectifier non-linearity $\rho$, 
and they are non-zero if $k$ is even \citep{mallat2020phase}. 
This result shows that for a suitable choice of $(\alpha,\alpha')$, 
the covariance statistics $C^\ALPHA x$ can implicitly capture 
$C^\WPH x$ with a wide range of $k$ and $k'$.

\paragraph{Relation with second order statistics}

Using a simple decomposition of wavelet coefficients into their positive, negative, real and imaginary parts, we can further show that the covariance statistics $ C^\ALPHA x $
capture the classical second order statistics of wavelet coefficients, 
also used in the PS model (with phase harmonic coefficients $k=k'=1$).

\begin{prop}\label{prop.second-order}
    Let $I=\{0,\frac{\pi}{2}, \pi, \frac{3\pi}{2}\}$.  
    There exists a finite complex-valued sequence $\{w_{\al, \al'}\}_{(\al, \al') \in I^2}$ such that for all $(j, \theta) \in \La$, and all $\tau \in \omN$,
    \begin{equation}\label{e.second-order}
        \sum_{ (\al, \al') \in I^2} w_{\al, \al'} C^\ALPHA x(( j,\theta,\al), ( j',\theta',\al'), \tau) = \sum_{u \in \omN} \big(x \star \psi_{j, \theta}(u)\big) \big(x \star \psi_{j', \theta'}(u - \tau)\big)^\ast.
    \end{equation}
\end{prop}

The proof is given in \Cref{proof-s-o}. 
This shows that using only four $\alpha$ uniformly chosen 
between $[0,2 \pi]$ is sufficient to capture second order statistics. 
Because the wavelet transform is an invertible linear operator (on its range space), computing the r.h.s of \cref{e.second-order} for all $(j, \theta, \tau)$, as well as the low-pass coefficients carried out by $\Phi_J$, is equivalent to computing the correlation matrix of $x$.







\paragraph{Relation with the RF model} Setting aside the subtraction by the spatial mean $\mu_\ga$, the RF model can be viewed as a particular case of models defined by \cref{alpha-features}. Indeed, the statistics of the RF model take the form of \cref{eq:corr}, with
\[
    R^{\RF}x (f, u) = \rho ( x \star \psi_{f} (u) ) ,
\]
where $\{\psi_{f}\}$ being a family of multi-scale random filters. By writing $\rho_\al(x \star \psi_{j, \theta}(u - \tau)) = \rho(x \star \mbox{Real}(\psi^\tau_{j, \theta} e^{i\al})(u))$, with $\psi^\tau_{j, \theta}$ denoting the translation of $\psi_{j, \theta}$ by $\tau$, we see that the models are similar, the difference being that our models use wavelet-based filters instead of random ones.

\subsection{Defining an appropriate $\U$}\label{ss.extremes}


The choice of the covariance set  $\U$ 
is of central importance in the definition of the model. 
Intuitively,
a too small set of indices will induce a model 
that could miss important structural information 
about the texture that we want to synthesize. 
Conversely, if $\U$ contains too many indices, 
the syntheses can have good visual quality, 
but the statistics of the model may have a large variance, 
leading to the memorization of some patterns of the observation. 
There is a trade-off between these two aspects: 
one must capture enough information to get syntheses 
of good visual quality, but not much, 
so as not to reproduce parts of the original image. 
To illustrate this point, we shall study the model ALPHA defined with three different sets $\U$ : A smaller model $\ALPHAS$ with a limited amount of elements in $\U$, an intermediate model $\ALPHAI$, 
and a larger model $\ALPHAL$. 

To precisely define these models, let us note $\mathfrak{J} := \{0, \cdots, J-1 \}$, $\Theta := \frac{\pi}{L}\{ 0, \cdots, L-1 \}$, and $\mathcal{A}_A = \frac{2\pi}{A}\{ 0, \cdots, A-1 \}$. Let us also define the set $\mathfrak{T} := \{0\} \cup \{ 2^j (\cos(\theta), \sin(\theta))\}_{0\leq j < J, \: \theta \in \frac{\pi}{L}\{ 0, \cdots, 2L-1 \} }$,
from which the spatial shift shall be selected.
Table \ref{tab:params} summarizes the conditions that all parameters have to satisfy to be contained in these sets. Additionally, these models include large scale information through the covariance of a low-pass filter, i.e. the spatial average of $ x \star \phi_J( \cdot) x \star \phi_J(\cdot - \tau)$, for $\tau \in \mathfrak{T}$. To count the size of $\U$ without redundancies, 
Appendix \ref{subsec:nbalpha} provides an upper bound on the non-redundant statistics in our models. 
This upper bound is used to count the number of statistics in our models. 
To keep this number from being too large, instead of taking all shifts in a square box, such as in \citet{portilla2000parametric}, we choose to select only shifts of dyadic moduli, and with the same orientations as the wavelets.

{\renewcommand{\arraystretch}{2}
\begin{table}[!h]
\caption{List of indices in $\U$ for different $\ALPHA$ models.}
    \small
    \centering
    \begin{tabular}{c|c|c|c|c | c }
      Model & Scales & Angles & Phase shift & Spatial shift & Size of  $\U$   \\
      \hline
      $\ALPHAS$ & $\substack{(j,j') \in \mathfrak{J}^2 \\ |j'-j| \leq 1}$ &  $(\theta, \theta') \in \Theta^2$ & ${\textstyle(\al, \al') \in \mathcal{A}_4 \! \times \! \mathcal{A}_1}$  & $\substack{\tau \in \mathfrak{T} \text{ if }(j, \theta) = (j', \theta') \\ \tau = 0 \text{ otherwise.}}$  &  ( $J |\Theta|^2 + J |\Theta|  |\mathfrak{T}| ) |\mathcal{A}_4| $ 
      \\
      \hline
      $\ALPHAI$ & $(j,j') \in \mathfrak{J}^2$ & $(\theta, \theta') \in \Theta^2$ & ${\textstyle(\al, \al') \in \mathcal{A}_4 \! \times \! \mathcal{A}_1}$  & $\tau \in \mathfrak{T}$ & $J^2 |\Theta|^2|\mathcal{A}_4| |\mathfrak{T}|$
      \\
      \hline
      $\ALPHAL$ & $(j,j') \in \mathfrak{J}^2$ &  $(\theta, \theta') \in \Theta^2$ & ${\textstyle(\al, \al') \in \mathcal{A}_4^2}$  & $\tau \in \mathfrak{T}$ & $J^2 |\Theta|^2|\mathcal{A}_4|^2 |\mathfrak{T}|$
    \end{tabular}
    \label{tab:params}
\end{table}
}

\paragraph{$\ALPHAS$ vs. $\ALPHAI$}
The small model $\ALPHAS$ is inspired from the PS model, 
as it only takes into account of the interactions 
between nearby scales (i.e. $|j'-j|\leq 1$), and 
the spatial shift correlations are only considered for $(j,\theta) = (j',\theta')$. 
There are two notable differences in the statistics included in the $\ALPHAS$ and $\ALPHAI$ models. The first one is the range of scales being correlated. It has been shown in \citet{zhang2021maximum} that constraining correlation between a wider range of scales induces a better model for non-Gaussian stationary processes, and a better estimation of cosmological parameters from observed data \citep{allys2020new}. The second difference, which has a significant impact on the number of statistics (it increases the model size by a factor $\sim$10), is the number of spatial shifts in the correlations. In the $\ALPHAI$ model, spatially shifted correlations are computed for all pairs of coefficient $(\ga, \ga')$. 
For both stationary textures and non-stationary images in gray-scale, shape and contours of salient structures and objects are better reproduced with $\ALPHAI$, as illustrated in Figure \ref{fig.small-big}. More examples are given in Appendix \ref{supp:alphasi}.


\paragraph{$\ALPHAI$ vs. $\ALPHAL$}
As we observe in \Cref{fig.small-big},
the $\ALPHAI$ model, containing 4 times less coefficients than the $\ALPHAL$, suffers less from memorization effects, while still capturing most of the geometric information in the images. 
This small loss of information can be 
partially explained
by the frequency transposition property of the phase harmonics operator \citep{mallat2020phase}, for compactly supported wavelets in the frequency domain, as detailed in \Cref{demo-alpha-prime}. In order to avoid this memorization effect, we shall, in the rest of the paper, consider only the intermediate model.







{\setlength{\tabcolsep}{2pt}
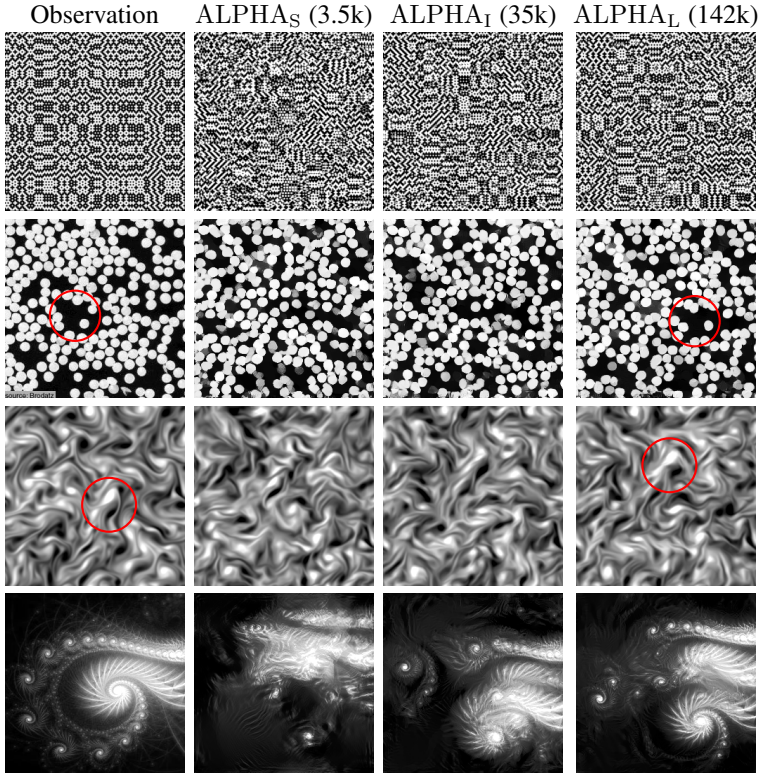
\begin{figure}[!h]
    \centering
    \begin{tabular}{cccc}
        Observation & ${\textstyle\ALPHAS}$ (3.5k) & ${\textstyle\ALPHAI}$ (35k) & ${\textstyle\ALPHAL}$ (142k) \\
         \includegraphics[width=0.17\linewidth]{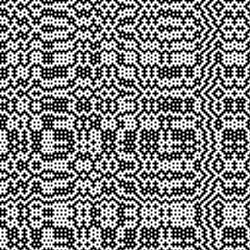}&
       \includegraphics[width=0.17\linewidth]{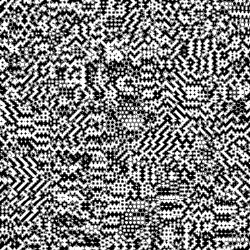}&
       \includegraphics[width=0.17\linewidth]{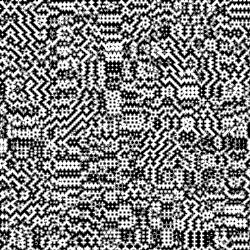} & 
       \includegraphics[width=0.17\linewidth]{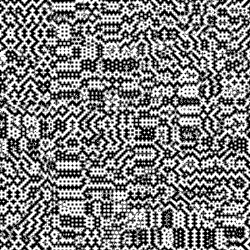}\\
       \begin{tikzpicture}
        \node[anchor=south west,inner sep=0] (image) at (0,0) {\includegraphics[width=0.17\textwidth]{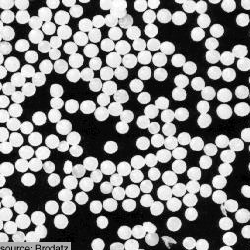}};
        \begin{scope}[x={(image.south east)},y={(image.north west)}]
            \draw[red, thick,] (0.39,0.46) circle (0.14);
    \end{scope}
    \end{tikzpicture}&
       \includegraphics[width=0.17\linewidth]{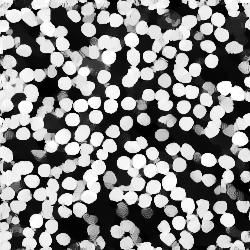}&
       \includegraphics[width=0.17\linewidth]{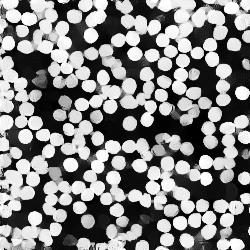}&
       \begin{tikzpicture}
        \node[anchor=south west,inner sep=0] (image) at (0,0) {\includegraphics[width=0.17\textwidth]{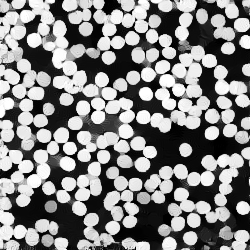}};
        \begin{scope}[x={(image.south east)},y={(image.north west)}]
            \draw[red, thick,] (0.66,0.43) circle (0.14);
    \end{scope}
    \end{tikzpicture}\\
         \begin{tikzpicture}
        \node[anchor=south west,inner sep=0] (image) at (0,0) {\includegraphics[width=0.17\textwidth]{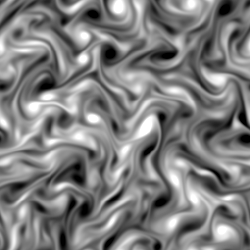}};
        \begin{scope}[x={(image.south east)},y={(image.north west)}]
            \draw[red, thick,] (0.58,0.45) circle (0.15);
    \end{scope}
    \end{tikzpicture}&
        \includegraphics[width=0.17\linewidth]{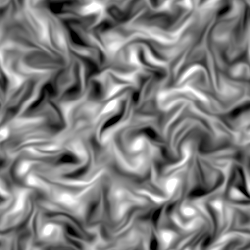}&
        \includegraphics[width=0.17\linewidth]{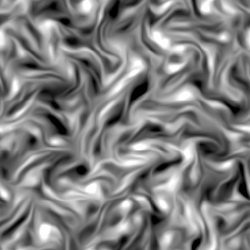}&
    \begin{tikzpicture}
        \node[anchor=south west,inner sep=0] (image) at (0,0) {\includegraphics[width=0.17\textwidth]{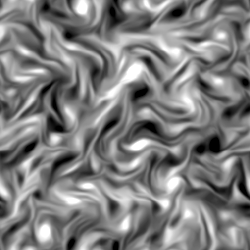}};
        \begin{scope}[x={(image.south east)},y={(image.north west)}]
            \draw[red, thick,] (0.52,0.67) circle (0.15);
    \end{scope}
    \end{tikzpicture}\\
    \includegraphics[width=0.17\linewidth]{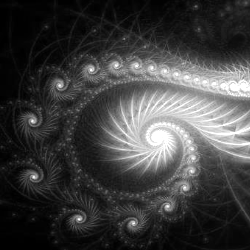}&
       \includegraphics[width=0.17\linewidth]{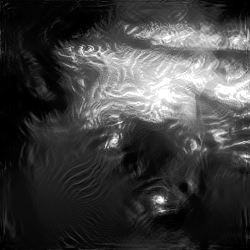}&
       \includegraphics[width=0.17\linewidth]{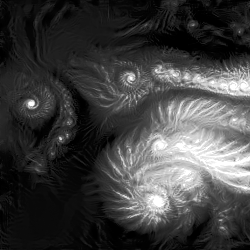} & 
       \includegraphics[width=0.17\linewidth]{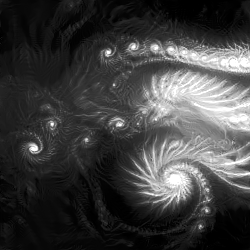}\\
    \end{tabular}
    \caption{Examples of syntheses from the $\ALPHA$ models defined in Table \ref{tab:params}. 
    Visually similar image patches in textures are highlighted by red circles.
    The number of statistics is given in brackets next to each model name.
    From top to bottom: a Julesz counterexample, a stationary texture image, a stationary turbulent field and a non-stationary image. 
    We see that the model $\ALPHAI$ achieves a balance between the visual quality and diversity on these examples. 
    }
    \label{fig.small-big}
\end{figure}}


\subsection{Modelling color interactions}
\label{sec:color}

In order to generate color textures, the covariance model $\ALPHAI$ defined in Section~\ref{ss.extremes} could be directly applied to each R, G and B color channel independently.
However, it would not take into account the color coherence in the structures of the observation.



%

To capture color interactions in the observation image, 
we shall impose the covariance between the coefficients 
of \cref{alpha-features} for all indices in $\U$ and all color channels.
More precisely, let $x = \{x^c\}_{c=1,2,3}$ be a color image, 
with the parameter $c$ representing the color channel. 
The $\ALPHAC$ color model is defined by correlations between 
coefficients of the form:
\begin{equation}
    R^{\ALPHAC} x(\ga, u) = \rho_\al(x^c \star \psi_{j, \theta})(u) - \mu_\ga, \quad \ga = (j, \theta, \al, c).
\end{equation}
The set of indices is defined as $\U^{\ALPHAC} := \{ (\ga, \ga', \tau) : ((j, \theta, \al), (j', \theta', \al'), \tau) \in \U^{\ALPHAI}, (c, c') \in \{1, 2, 3\}^2 \}$. 

\paragraph{Reduced $\ALPHAC$}
The model $\ALPHAC$ has a large size as it computes correlations between all coefficients for all color channels. This size can be significantly reduced by computing spatially shifted coefficients only for the same color channels (to capture their geometries). This reduced model contains three times less coefficients ($\sim$113k), without significant reduction of the visual quality of syntheses, as detailed in \Cref{app.reduced-c}.

\section{Numerical results}
\label{s.results}

In this section, 
we compare our intermediate model 
to the state-of-art models (PS, RF and VGG)
on both gray-scale and color textures. 
We first specify the experimental setup. 
We then present the synthesis results of various examples, and discuss their quality 
through visual inspection.
A quantitative evaluation of the quality of the syntheses, based on the synthesis loss of the VGG model, and proposed in \citet{ustyuzhaninov2017what}, 
is discussed in Appendix \ref{app:vgg}. 








\subsection{Experimental Setup}\label{ss.setup}

For our experiments, we choose gray-scale and color textures with various visual structures.\footnote{The source for the presented textures are given in Appendix \ref{app:params}.}
In the gray-scale examples, we also include a stationary turbulent field (vorticity), 
which is simulated from Navier-Stokes equations in two dimensions \citep{schneider2006coherent}. 
These examples all contain complex geometric structures 
that are hard to model by the classical PS model. 

The texture images presented in this work have a size of $N=256$, giving a number of pixels of $\sim$65k.
For all the $\ALPHA$ models, 
we use Morlet wavelets with a maximal scale $J=5$ and number of orientations $L=4$. 
This choice differs from the PS model, 
which uses Simoncelli steerable wavelets.
In Appendix \ref{app:wavelet}, 
we discuss the impact of these wavelets on our model.
To draw the samples, we follow 
gradient-based sampling algorithms, suitable in high-dimensions \citep{bruna2018multiscale}, to minimize the objective function $\| Cx - C\bar{x}\|^2$, starting from an initial sample from a normal distribution.
Similarly to \citet{gatys2015texture, ustyuzhaninov2017what}, we use the L-BFGS algorithm \citep{nocedal1980updating} for the optimization of the objective.
As in the VGG model \citep{gatys2015texture}, 
we further apply a histogram matching procedure as post-processing after 
optimization. The details of the PS, RF and VGG models, 
as well as detailed specifications of our models, are given
in Appendix \ref{app:params}. 
More synthesis examples can be found in \Cref{supp:comp}. 

\subsection{Results}\label{ss.results}




In \Cref{fig:comp-ps-vgg}, 
we present examples of syntheses from 
the $\ALPHAI$ (or $\ALPHAC$), PS, RF and VGG models,
for both gray-scale and color textures, as well as for non-stationary images. 
We observe that our model $\ALPHAI$ produces texture syntheses of 
similar visual quality to the RF and VGG models. 
It also significantly outperforms the PS model in terms of the visual quality, 
without introducing visible memorization effects.
As the model PS uses the statistics closer to $\ALPHAS$ compared to $\ALPHAI$, 
the performance of PS is somehow expected. 


Note that for the tiles example (the fifth row) in \Cref{fig:comp-ps-vgg}, 
the VGG model produces less convincing textures, 
because the long-range correlations present in the image (aligned tiles) are not reproduced. 
To remedy this issue, it has been proposed in \citet{berger17incorp} to add spatial shifts to the correlations of the network feature maps. 
These shift statistics are similar to the parameter $\tau$ in our model. 
We also observe that, in the case of the sixth row example (flowers), all models fail to reproduce complex structures at object-level. Possible improvements of such models is further discussed in \Cref{s.discussion}.

For non-stationary images, 
we find that certain image patches can be more or less memorized 
by the RF, VGG and $\ALPHAI$ models, as illustrated in the seventh row example. 
Understanding such memorization effect of non-stationary images is a subtle topic, 
as we find that in some binary images ($\bar{x}(u) \in \{0,1\}$), only the
PS model can reproduce the observation, even though it has a much smaller number of statistics (the last row example).
We find that this is related to the spatial correlation statistics in PS (non-zero $\tau$). 
By removing this constraint,
$\bar{x}$ is no longer always reproduced.\footnote{Set the parameter $\Delta=0$ in the PS model. See Appendix \ref{app:params} for more details.
This simple example suggests that sometimes it is very important to choose the 
right statistics to capture specific geometric structures. 
}
The non-stationary nature also appear in some logo near the boundary of some textures (e.g. bottom left in the observation of the first and fourth rows). 
Although this logo 
is reproduced by RF, VGG, $\ALPHAI$ and $\ALPHAC$, it is a very local phenomenon,
as we do not find visible copies of the textures when there is no logo, and it 
is likely related to the way one addresses the boundary effect (see more in Appendix \ref{app:nonper}).







{\setlength{\tabcolsep}{2pt}
\begin{figure}[!h]
    \centering
    \begin{tabular}{ccccc}
        Observation & PS (3.2k/17k) & RF (525k) & Ours (35k/320k) & VGG (177k) \\
        \includegraphics[width=0.17\linewidth]{./figs/obs/cerise_original2.pdf}&
        \includegraphics[width=0.17\linewidth]{figs/ps/cerise_ps_gray_J5L8Dn4.pdf}&
        \includegraphics[width=0.17\linewidth]{figs/rf/cerise_rf_gray_nobias_nit2000_size256.pdf}&
        \includegraphics[width=0.17\linewidth]{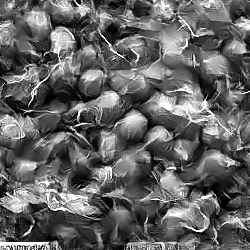}&
        \includegraphics[width=0.17\linewidth]{./figs/vgg/cerise_vgg.pdf}\\
         \includegraphics[width=0.17\linewidth]{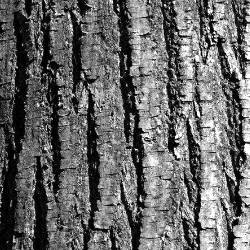}&
        \includegraphics[width=0.17\linewidth]{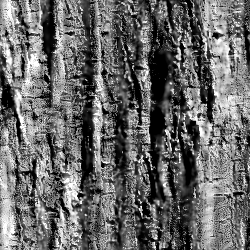}&
        \includegraphics[width=0.17\linewidth]{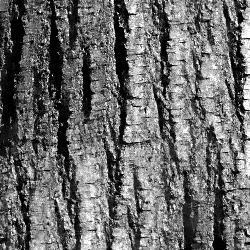}&
        \includegraphics[width=0.17\linewidth]{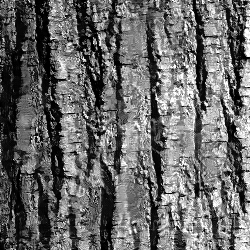}&
         \includegraphics[width=0.17\linewidth]{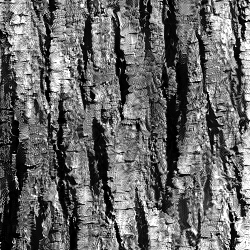}\\
        \includegraphics[width=0.17\linewidth]{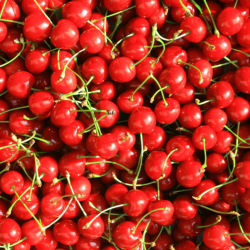}&
        \includegraphics[width=0.17\linewidth]{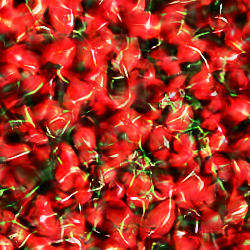}&
        \includegraphics[width=0.17\linewidth]{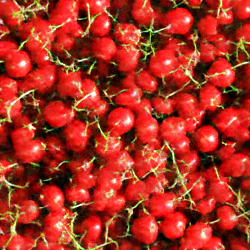}&
        \includegraphics[width=0.17\linewidth]{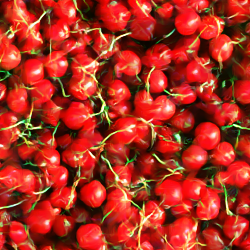}&
        \includegraphics[width=0.17\linewidth]{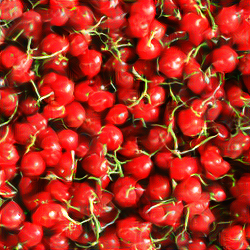}\\

        \includegraphics[width=0.17\linewidth]{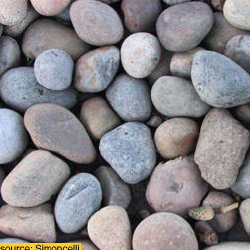}&
        \includegraphics[width=0.17\linewidth]{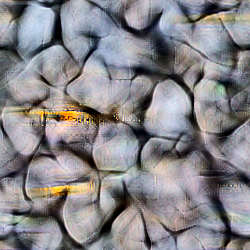}&
        \includegraphics[width=0.17\linewidth]{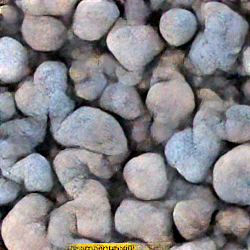}&
        \includegraphics[width=0.17\linewidth]{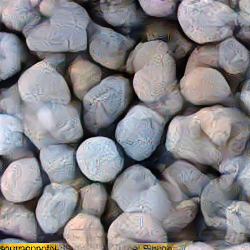}&
        \includegraphics[width=0.17\linewidth]{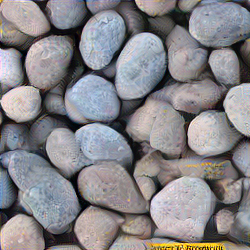}\\
        \includegraphics[width=0.17\linewidth]{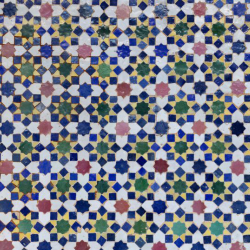}&
         \includegraphics[width=0.17\linewidth]{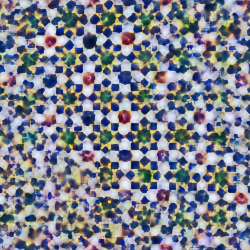}&
         \includegraphics[width=0.17\linewidth]{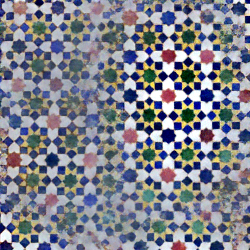}&
         \includegraphics[width=0.17\linewidth]{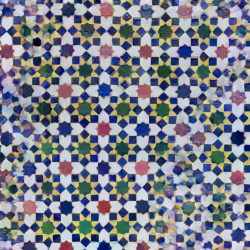}&
         \includegraphics[width=0.17\linewidth]{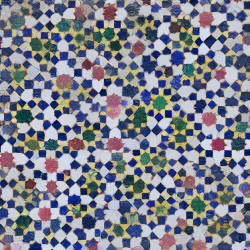}\\
        \includegraphics[width=0.17\linewidth]{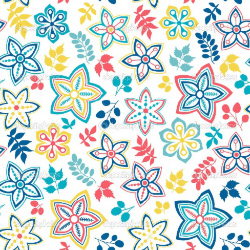}&
        \includegraphics[width=0.17\linewidth]{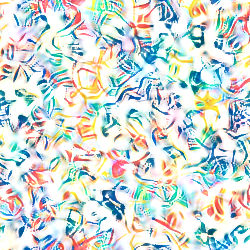}&
        \includegraphics[width=0.17\linewidth]{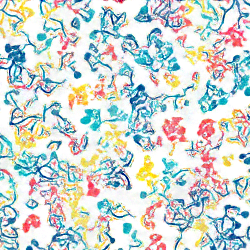}&
        \includegraphics[width=0.17\linewidth]{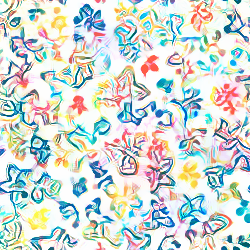}&
        \includegraphics[width=0.17\linewidth]{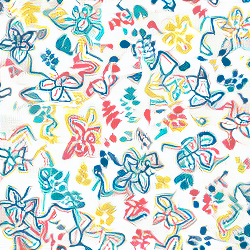}\\
        \includegraphics[width=0.17\linewidth]{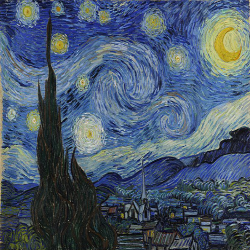}&
         \includegraphics[width=0.17\linewidth]{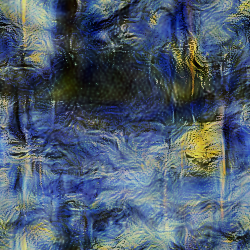}&
         \includegraphics[width=0.17\linewidth]{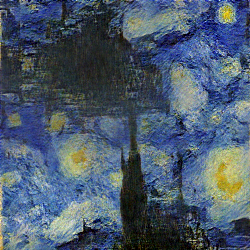}&
         \includegraphics[width=0.17\linewidth]{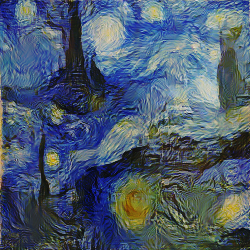}&
         \includegraphics[width=0.17\linewidth]{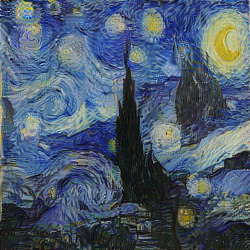}\\
        \includegraphics[width=0.17\linewidth]{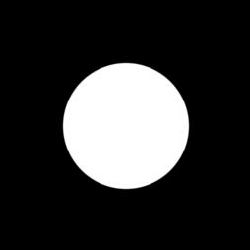}&
        \includegraphics[width=0.17\linewidth]{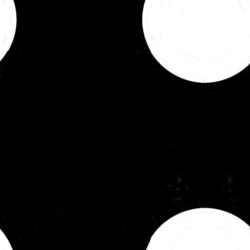}&
        \includegraphics[width=0.17\linewidth]{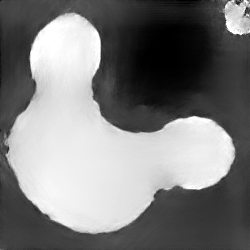}&
        \includegraphics[width=0.17\linewidth]{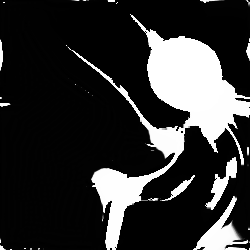}&
        \includegraphics[width=0.17\linewidth]{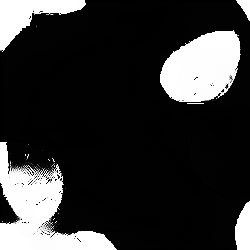}\\
    \end{tabular}
    \caption{Visual comparison between the gray-scale and color models. 'Ours' denotes the $\ALPHAI$ model for gray-scale images and the $\ALPHAC$ model for color images. 
    }
    \label{fig:comp-ps-vgg}
\end{figure}}

\section{Discussion}\label{s.discussion}



In this work, we presented a new wavelet-based model for natural textures. This model incorporates a wide range of statistics, by computing covariances between rectified wavelet coefficients, at different scales, phases and positions. We showed that this model is able to capture and reproduce complex geometric structures, present in natural textures or physical processes, producing syntheses of similar quality to state-of-the-art models that use CNN-based representations.

Being defined with a wavelet family instead of multi-scale random filters, the proposed model uses less statistics than the RF model.
For the gray-scale textures, our model has about 15 times less statistics, as it focuses on capturing the geometric structures present in images. 
Although our color model has a slightly larger number of statistics than VGG, 
the reduced color model, presented in Section \ref{sec:color}
is three times smaller than $\ALPHAC$, while achieving similar visual quality. 
It shows the potential 
to further reduce the size of the color model. 
In the PS model, a PCA on the color channels is performed \citep{vacher2021portilla}. The same idea could also be applied to our model. 

Furthermore, there are examples where all the models may all fail to produce some geometric structures at object-level, as illustrated in \Cref{fig:comp-ps-vgg} (sixth row). 
In this situation, we still need to find more informative statistics. 
One may for example consider to incorporate a second layer of wavelet transform as in the wavelet scattering transform \citep{leonarduzzi2019maximum, he2021texture}. 
Another line of research is to introduce other kinds of losses
(such as to encourage image smoothness) in order to improve the VGG model 
\citep{liu2016texture,sendik2017deep}. 
These losses are complementary, and could thus also be added to our models. 
Integrating these models with learning-based approaches is another promising direction
\citep{zhou2018non,zhu1997minimax,xie2016theory,xie2018cooperative}.

Finding a minimal set of statistics to define a texture model remains important because a large number of statistics can result in a high variance of the estimators, and the associated model
may suffer from memorization effects.
This is a problem because the aim of the model is to approximate the underlying distribution of the observation, and therefore produce diverse textures. In this regard, the mere visual evaluation of the synthetic textures
can fail to take this aspect of the model into account. 
Defining a quantitative evaluation of quality \emph{and} diversity, 
coherent with visual perception, 
remains an open problem
\citep{ustyuzhaninov2017what,yu2019texture}.

\paragraph{Acknowledgments}
We thank all the reviewers for their insightful feedback.
This work was partially supported by a grant from the PRAIRIE 3IA Institute of the French ANR-19-P3IA-
0001 program.
Sixin Zhang acknowledges the support provided by 3IA Artificial and Natural Intelligence
Toulouse Institute, French "Investing for the Future - PIA3" program
under the Grant agreement ANR-19-PI3A-0004, and by 
Toulouse INP ETI and AAP unique CNRS 2022 under the Project LRGMD.


\paragraph{Reproducibility Statement} 
In order to reproduce the experiments presented in this work, the main parameters to define the model can be found in Section \ref{ss.setup}. 
More details are given in Appendix \ref{app:params}. 
Additionally, the code is made publicly available.

\paragraph{Ethics Statement} The authors acknowledge that no potential conflicts of interest, discrimination, bias, fairness concerns or research integrity issues have been raised during the completion of this work.

\bibliography{refs}
\bibliographystyle{iclr2022_conference}

\appendix

\section{Model and algorithmic specification}\label{app:params}

We provide additional information needed to reproduce the numerical results 
of the models (for both gray-scale and color textures) 
considered in this paper. 
First, we detail the models of PS, RF, VGG.  
We then give algorithmic parameters 
to obtain the synthesis images.
For natural textures, we also propose a strategy to synthesize 
non-periodic images in our models. An image is non-periodic 
if a periodic extension of the image to the domain outside $\omN$ 
create discontinuities at the contour of $\omN$. 

\paragraph{Sources of textures}
Our natural texture examples were obtained from the following three sources: CNS NYU\footnote{\url{http://www.cns.nyu.edu/~lcv/texture/}},
Textures.com\footnote{\url{https://textures.com/}}, Describable Textures Dataset model\footnote{\url{https://www.robots.ox.ac.uk/~vgg/data/dtd/index.html}} and the Github page of \cite{berger17incorp}
\footnote{ \url{https://github.com/guillaumebrg/texture_generation}}.

\subsection{Model parameters}

We specify the model parameters 
to synthesize both gray-scale and color textures. 
We also discuss how to extend 
the RF and the VGG model, 
originally designed
for color textures, 
to model gray-scale textures. 

\begin{itemize}

\item PS: For both gray-scale and color models, 
we set the number of scales $J=5$, 
and the number of orientations $L=8$ 
for the Simoncelli steerable wavelets. 
The spatial shift $\tau=(\tau_1,\tau_2)\in \omN$ is 
chosen to be in the range of 
$\max{(\tau_1,\tau_2)} \leq \Delta=4$. 
Note that this is different to the model parameters 
reported for the default PS model (which is $J=4,L=4,\Delta=3$)
as we find that it results in 
a larger set of statistics and better visual quality. 
The synthesis results of this model can be reproduced by a Matlab software.\footnote{\url{https://www.cns.nyu.edu/~lcv/texture/}}

\item RF: 
For gray-scale textures, we consider $J \times L$
random convolutional filters $(\psi_{f})_{1 \leq f \leq J L}$.
Let $f = (j,\ell)$, 
the index $j $ representing the scale of each filter, 
whose size is $(W_j,W_j)$ for $ 1 \leq j \leq J$.
For $ \Ga = \{  f = (j,\ell) | j \leq J, \ell \leq L \} $, 
the representation is 
\[
    R^{\RF} x  (\ga, u) = \rho  ( x \star \psi_{j,\ell} (u) ) , \quad \ga  = (j,\ell)  \in \Ga.
\]

For a color image $x = \{ x^c  \}_{ 1 \leq c \leq 3}$, 
we use $3 \times J \times L$ random convolutional filters. 
The representation of $x$ is
\[
    R^\RF x  (\ga, u) =
    \rho \big( \sum_{c=1}^3  x^c  \star \psi_{c,j,\ell} (u) \big), 
    \quad \ga  = (j,\ell) \in \Ga.
\]
The correlations $C^{\RF} x$ 
are defined for all pairs $(\ga,\ga') \in \Ga \times \Ga$.
In both the gray and color cases, it results 
in a correlation matrix 
$C^\RF$ with $J^2 L^2 $ statistics. 

Following the default setting of the RF model, 
we set $J=8$ and  $L=128$ for filters whose sizes are
$W_1 = 3, W_2 = 5, W_3 = 7, W_4 = 11, W_5 = 15, W_6 = 23, W_7 = 37, W_8 = 55$. 
Each filter $\psi_{j,\ell}$ or $\psi_{c,j,\ell}$ is generated randomly 
according to \url{GlorotUniform} in the software Lasagne.
\footnote{\url{https://lasagne.readthedocs.io/}}

\item VGG. 
For a color image $x = \{ x^c \}$, 
the VGG model computes a correlation matrix 
$C^{\VGG} x$ between the features maps 
within different layers of a pre-trained CNN network. 
To adapt this model to gray-scale textures, 
we shall add one input layer which converts 
a gray-scale image $y$ into a color image, by 
setting $x^c = y$ for each color channel $c$. 
This allows one to use the same $C^{\VGG} x$ to compute 
the gradient of the VGG loss with respect to $y$,
and therefore to synthesise a gray-scale texture. 
For both gray-scale and color textures, 
we use only five layers 
'conv1\_1', 'pool1', 'pool2', 'pool3', 'pool4',
as proposed in the original work. 

\item $\ALPHA$. See the main text.  

\end{itemize}

\subsection{Algorithmic parameters}

We specify the optimization parameters used
to synthesize both gray-scale and color textures. 

\begin{itemize}

\item PS: It utilizes iterative projections 
onto constraint sets to generate textures. 
We set the number of iterations to 200. 

\item RF: It uses the L-BFGS procedure\footnote{\url{scipy.optimize.fmin_l_bfgs_b} in Python}
with a memory size 20, and with a maximal number of iterations 2000.
The initialization for each pixel value of 
a gray-image is Uniform between $[-1,1]$.
For the color image case,  
each RGB channel 
is initialized independently
with a Uniform distribution between $[-1,1]$. 
To address non-zero mean textures (i.e. $\E(X(u)) \neq 0$), 
the empirical mean of $\bar{x}$ is subtracted 
from the input $x$ to compute the representation. 
It is added back to the output of the optimization to produce a synthesis. 



\item VGG: 
It uses the L-BFGS procedure\footnote{\url{scipy.optimize.minimize}  in Python}
with a memory size of 20, 
and with a maximal number of iterations of 2000.For both gray-scale and color images, bounds constraints are used for the optimization.
The initialization of each pixel value 
is the standard normal distribution (zero mean, unit variance). 
To address non-zero mean textures (i.e. $\E(X(u)) \neq 0$), 
the VGG mean is subtracted 
from the input $x$ of the representation\footnote{For the color image whose pixel value 
is between zero and one, 
the mean of BGR is 0.40760392,  0.45795686,  0.48501961. For the gray-scale, we simply take the average 
of the BGR mean.}. 
It is added back to the output after the optimization to produce a synthesis 
(with an additional histogram matching post-processing). 


\item $\ALPHA$: 
For all the models in $\ALPHA$, 
we use the L-BFGS optimization 
algorithm with restarts. 
Starting from the standard normal distribution, with mean and standard deviation estimated from the observation, 
we use the L-BFGS procedure 
implemented in Pytorch. 
It runs for 500 iterations and then 
it is restarted with an initialization 
obtained from the previous L-BFGS result. 
This is repeated 10 times to obtain the synthesis
(with an additional histogram matching post-processing). 


\end{itemize}

\subsection{Number of statistics in the $\ALPHA$ models}
\label{subsec:nbalpha}
In this section, we detail the number of statistics in the different $\ALPHA$ models. We begin by giving the formula for each model (note that we do not include the low-pass statistics, which numbers are negligible). To (partially) avoid redundancy in the coefficients, for all models, we compute only the correlations for indices in $\U$ such that $j_2 \geq j_1$. This gives us the following formulas for the number of statistics:

\begin{itemize}
    \item $\#(\ALPHA_S) = (2J-1)|\Theta_4|^2|\mathcal{A}_4| + J|\Theta_4||\mathcal{A}_4||\mathfrak{T}|$
    \item $\#(\ALPHAI) = \frac{1}{2}J(J+1)|\Theta_4|^2|\mathcal{A}_4||\mathfrak{T}|$
    \item $\#(\ALPHAL) =\frac{1}{2}J(J+1)|\Theta_4|^2|\mathcal{A}_4|^2|\mathfrak{T}|$
    \item$\#(\ALPHAC) =\frac{9}{2}J(J+1)|\Theta_4|^2|\mathcal{A}_4||\mathfrak{T}|$
\end{itemize}

Note that, for all $\ALPHA$ models, we also compute first order statistics $\mu_\ga$, i.e. the spatial averages of $R^\ALPHA \bar{x}(\ga, u)$. There are $J|\Theta_4||\mathcal{A}_4|$ statistics of this sort in every model, which is negligible with respect to the total number of second order statistics.

Note also that there are still some redundancies in these statistics, as for $j=j'$, all correlations for $\theta, \theta', \al, \al', \tau$ are counted twice. 
The number of such statistics is\footnote{The number of such moments is of the order of $J|\Theta|^2|\mathcal{A}_4|$ for the small model, $J|\Theta|^2|\mathcal{A}_4|^2$ for the intermediate and large models, and $9J|\Theta|^2|\mathcal{A}_4|^2$ for the color model.}, for all model, superior to the number of first order statistics, which shows that our formula is in fact an upper bound for the exact number of statistics.

\subsection{Number of statistics in the PS models}\label{app:nbps}

The PS model can be interpreted as a particular case of the WPH covariance model as it contains the following three key categories of statistics: 

$\bullet$ Raw coefficient correlations ($k=k'=1$): they capture 2nd order statistics of a stationary process $X$, i.e. the correlations between $X(u)$ and $X(u')$ for $(u,u') \in \omN^2$. 

$\bullet$ Coefficient magnitude statistics ($k=k'=0$): they capture information of $X$ beyond the 2nd order statistics. Very often, nearby scales and angles are considered in the model, such as $j' \approx j$ and $\theta' \approx \theta$.

$\bullet$ Cross-scale phase statistics ($k=1,k'=2$): they capture local phase alignments of the wavelet coefficients at nearby scales $j'=j+1$, which are complementary to the magnitude correlations. 

The major issue to count the number of statistics of this model is to avoid double counting the statistics which are the same. This is mostly due to the symmetries of the covariance matrices. We next detail how obtain the number 
of statistics for both the gray-scale and color model, 
by following the work of \cite{portilla2000parametric} and \cite{vacher2021portilla}. 

As before, we assume the number of wavelet scales is $J$, the number of wavelet orientations is $L$, and the spatial shift range is within a square of size  $(2\Delta+1 ) \times (2\Delta+1)$. Let us denote $N_a = 2\Delta+1$. 
We next describe in detail the number of statistics of each category counted in our paper, 



\subsubsection{PS in gray-scale}

$\bullet$ Marginal statistics of $x$: $6$. They include mean, variance, skewness, etc. 

$\bullet$ Marginal statistics of wavelet coefficients: $2 (J+1) + 1$. 

$\bullet$ Auto-correlation of wavelet coefficients (raw coefficient correlations): $(J+1) (N_a^2 + 1) / 2$.

$\bullet$ Auto-correlation of magnitude of wavelet coefficients (coefficient magnitude statistics): $J L (N_a^2 + 1) / 2 + JL(L-1)/2 + (J-1) L^2 $.

$\bullet$ Mean of magnitude of wavelet coefficients: $JL + 2$. This is not counted in the paper of \cite{portilla2000parametric}, but it used in the Matlab software.

$\bullet$ Cross-correlation of phase of wavelet coefficients (cross-scale phase statistics): $2 (J-1) L^2 + J L^2 $. 
The extra $J L^2 $ coefficients are cross-correlation of real sub-band cousin coefficients, which are not counted in the paper, but used in the Matlab software. 

To compare with the model size of the original work of \cite{portilla2000parametric},
one can check that the sum of these number is 792 when $J=4,L=4,\Delta=3$ ($N_a=7)$. If we do not count the coefficients which are only counted in the Matlab software ($JL + 2 + J L^2  = 82$), it results in 710 as reported in the original paper.

\subsubsection{PS in color}

$\bullet$ Marginal statistics of $x$ and PCA transform of $x$: $6 \times 3 + 3 \times 4 = 30$. 

$\bullet$ Marginal statistics of wavelet coefficients: $ 6(J+1) + 9$. 

$\bullet$ Auto-correlation of wavelet coefficients (raw coefficient correlations): $3 (J+2) (N_a^2 + 1) / 2$. This is called Central autoCorr of the PCA bands in the Matalb software (which includes lowband).

$\bullet$ Auto-correlation of magnitude of wavelet coefficients (coefficient magnitude statistics): $3 J L (N_a^2 + 1) / 2 +  
J (3 L) (3L-1)/2 + (J-1) (3L)^2 $.

$\bullet$ Mean of magnitude of wavelet coefficients: $3 (JL + 2)$.

$\bullet$ Cross-correlation of phase of wavelet coefficients (cross-scale phase statistics): $J (3L)^2 + (J-1) (3L)(6L) $. 


\subsection{Non periodic boundaries in natural images}\label{app:nonper}

The convolution operation in the wavelet transform 
(\eqref{periodic}) is performed using the Fast Fourier Transform. Additionally, recall from \Cref{ss.framework} that spatial shifts are defined with periodic boundary conditions.
This implies periodicity of the input image $x$. 
However, natural texture images are not periodic, 
so one needs to adapt the computation 
of coefficients to take into account 
possible border effects. 
To that end, instead of averaging over all 
$u \in \Omega_N$ as in \cref{eq:corr}, 
each correlation coefficient is averaged 
over a sub-window inside $\Omega_N$, 
which size depends on the scales of 
the coefficients being correlated.
More precisely,
let $\ga = (j, \theta, \al)$ and $\ga' = (j', \theta', \al')$. 
Note $j_m := \max(j,j')$. We define $\Omega_{j_m} := \{ u=(u_1, u_2) \in \Omega_N : \: 2^{j_m} \leq u_i < N-2^{j_m},\: i= 1,2 \}$. Then, for non periodic images, we compute

\begin{equation}
    C^\ALPHA x(\ga, \ga', \tau) =\frac{1}{|\Omega_{j_m}|} \sum_{u \in \omN} \mathbbm1_{\Omega_{j_m}}(u)\mathbbm1_{\Omega_{j_m}}(u - \tau) R^\ALPHA x(\ga, u) R^\ALPHA x(\ga', u - \tau),
\end{equation}

where the spatial shifts are defined periodically. Note that the spatial averages $\mu_\ga$ and $\mu_{\ga'}$ are also performed on $\Omega_{j_m}$.

\section{Proof of \Cref{prop.equiv}}\label{proof-equiv}


Using the fact that 
\begin{align*}
    \rho(\mbox{Real}(z e^{i\al})) &= \rho(\mbox{Real}(|z|e^{i(\varphi(z)+\al)}))\\
    &=|z|\rho(\cos(\al + \varphi(z)),
\end{align*}

and computing the Fourier coefficients of the $2\pi$-periodic function $\rho_\al(z)$ in the variable $\al$, we obtain
\begin{align*}
    \F({\rho}_\al(z))(k) &:= \frac{1}{2\pi}\int_{[0, 2\pi]}\rho_\al(z)e^{-ik\al}d\al\\
    &= |z| \frac{1}{2\pi}\int_{[0, 2\pi]} \rho(\cos(\al + \varphi(z))e^{-ik\al}d\al\\
    &= |z| e^{ik\varphi(z)} c_k\\
    &= [z]^k c_k,
\end{align*}

where $c_k$ is the Fourier transform of $h(.) := \rho(\cos(.))$ at the frequency $k$. The function $\al \mapsto \rho_\al(z)$ being periodic in $\al$, we have its decomposition in Fourier series
\begin{align*}
    \rho_\al(z) &= \sum_{k \in \Z} \F({\rho}_\al(z))(k) e^{ik\al}\\
    &= \sum_{k \in \Z} c_k [z]^k e^{ik\al}.
\end{align*}

We can then write, for any $z, z' \in \C$, and $\al, \al' \in [0, 2\pi]$,
\[
    \rho_\al(z) \rho_{\al'}(z')^\ast = \sum_{k,k' \in \Z^2} c_k c_{k'}^\ast [z]^k [z']^{-k'} e^{i(k\al - k'\al')}.
\]

Replacing $z$ and $z'$ by any two wavelet coefficients $x \star \psi_{j, \theta} (u)$ and $x \star \psi_{j', \theta'} (u -\tau )$, we thus obtain the relation in \Cref{prop.equiv}.

\section{Proof of \Cref{prop.second-order}}\label{proof-s-o}


Let $z \in \C$, and recall from \cref{alpha-relu}, that $\rho_\al(z) = \rho(\mbox{Real}(z e^{i\al}))$.
Note that we have the following relation
\begin{equation}
\tag{9}
    z = \rho_0(z) - \rho_\pi(z) - i(\rho_{\frac{\pi}{2}}(z) - \rho_{\frac{3\pi}{2}}(z)).
\end{equation}

We can then write
\begin{align*}
    zz'^\ast&= \big(\rho_0(z) - \rho_\pi(z) - i(\rho_{\frac{\pi}{2}}(z) - \rho_{\frac{3\pi}{2}}(z))\big) \big(\rho_0(z') - \rho_\pi(z') - i(\rho_{\frac{\pi}{2}}(z') - \rho_{\frac{3\pi}{2}}(z'))\big) \\
    &= \sum_{\al, \al' \in I^2} w'_{\al, \al'} \rho_\al(z)\rho_{\al'}(z'),
\end{align*}

with $ I = \{ 0, \frac{\pi}{2}, \pi, \frac{3\pi}{2} \}$. Replacing $z$ with $x \star \psi_{j, \theta}(u)$, $z'$ with $x \star \psi_{j', \theta'}(u - \tau)$,, and injecting this relation in \cref{eq:corr} gives us the desired result, with $w_{\al, \al'} = \omN w'_{\al, \al'}$.

\section{Supplementary results for the gray-scale $\ALPHA$ models}
\label{supp:alphasi}

Here, we present further visual comparison between the different $\ALPHA$ models for gray-scale images, to illustrate the trade-off between quality and diversity.

{\setlength{\tabcolsep}{2pt}
\begin{figure}[!h]
    \centering
    \begin{tabular}{cccc}
       Obs & $\ALPHAS$ (3.5k) & $\ALPHAI$ (35k) & $\ALPHAL$ (142k)\\
          \includegraphics[width=0.2\linewidth]{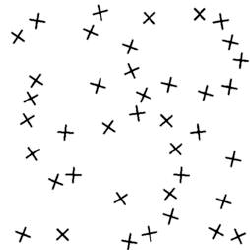}&
      \includegraphics[width=0.2\linewidth]{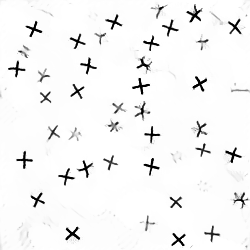}&
      \includegraphics[width=0.2\linewidth]{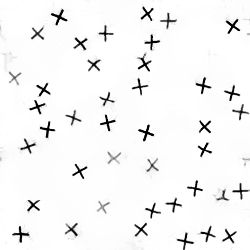}&
      \includegraphics[width=0.2\linewidth]{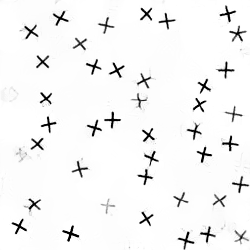}\\
       \includegraphics[width=0.2\linewidth]{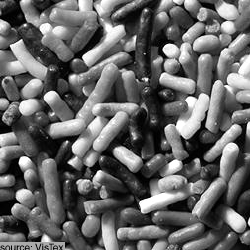}&
       \includegraphics[width=0.2\linewidth]{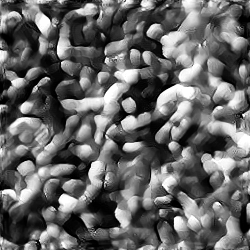}&
       \includegraphics[width=0.2\linewidth]{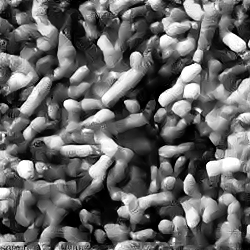} &
       \includegraphics[width=0.2\linewidth]{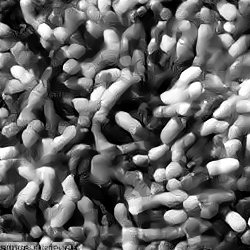}\\
       \includegraphics[width=0.2\linewidth]{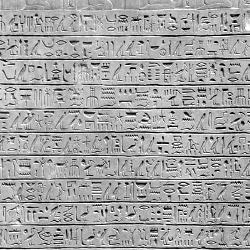}&
       \includegraphics[width=0.2\linewidth]{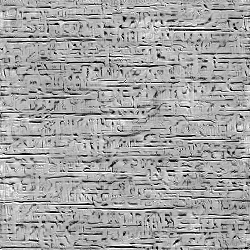}&
       \includegraphics[width=0.2\linewidth]{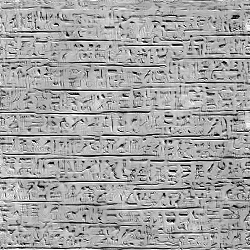} & 
       \includegraphics[width=0.2\linewidth]{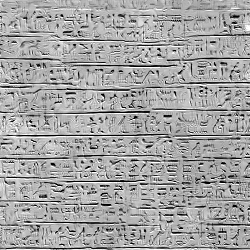}\\
       \includegraphics[width=0.2\linewidth]{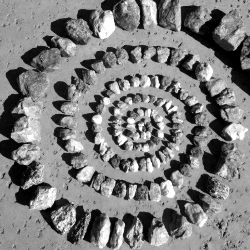}&
       \includegraphics[width=0.2\linewidth]{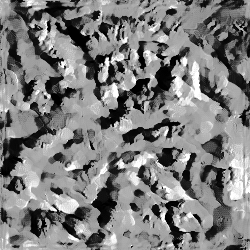}&
       \includegraphics[width=0.2\linewidth]{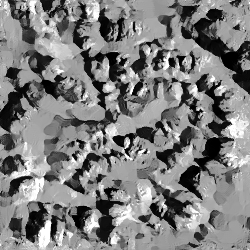}&
       \includegraphics[width=0.2\linewidth]{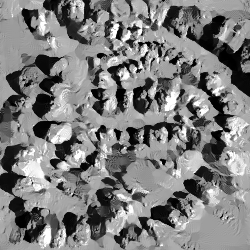}\\
       \includegraphics[width=0.2\linewidth]{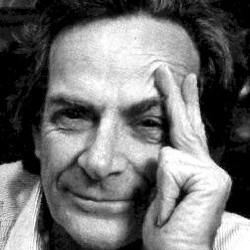}&
       \includegraphics[width=0.2\linewidth]{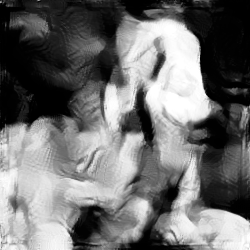}&
       \includegraphics[width=0.2\linewidth]{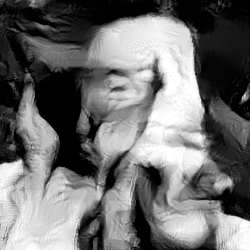}&
       \includegraphics[width=0.2\linewidth]{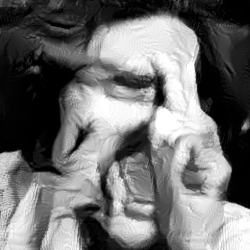}\\
           \end{tabular}
    \caption{Visual comparison of syntheses from the $\ALPHAS$, $\ALPHAI$ and $\ALPHAL$ models.}
    \label{fig:supp-comp-alpha}
\end{figure}
}

\section{Relation between $\ALPHAL$ and $\ALPHAI$}\label{demo-alpha-prime}
Here, we informally explain why, under some conditions on the wavelet family,
setting $\al' \in \{0\}$ in the $\ALPHA$ models should not lose too much (but still some) information captured by the statistics.

First, remark that the simple linear relation
\begin{equation}\label{e.lin}
\big( \sum_{\al \in \mathcal{A}_4} \rho_\al(z) e^{ik\al} \big) \rho_0(z) = \sum_{\al \in \mathcal{A}_4} \rho_\al(z) \rho_0(z) e^{ik\al}
\end{equation}
tells us that computing all correlations for $\al \in \mathcal A_4$ and $\al' = 0$ gives us at least the information contained in the r.h.s. of \cref{e.lin}.

Furthermore, if $\al \in \mathcal{A}_4 = \{ 0, \cdots, \frac{3\pi}{4} \}$, we make the following approximation
\[
\sum_{\al \in \mathcal{A}_4} \rho_\al(z) e^{ik\al} \simeq \int_{[0, 2\pi]} \rho_\al(z)e^{-ik\al}d\al = 2\pi c_k [z]^k.
\]

Recall also from the proof of \Cref{prop.second-order}, that $\F({\rho}_\al(z))(k) = [z]^k c_k$, where the Fourier transform is taken along the variable $\al$. Therefore,

\[
    \rho_0(z) = \sum_{k \in \Z} c_k [z]^k.
\]

Therefore,
\begin{align*}
    \big( \sum_{\al \in \mathcal{A}_4} \rho_\al(z) e^{ik\al} \big) \rho_0(z) & \simeq \big( \int_{[0,2\pi]} \rho_\al(z)e^{-ik\al}d\al \big) \rho_0(z) \\
    &= 2\pi c_k[z]^k \big( \sum_{k' \in \Z} c_{k'} [z']^{k'} \big)\\
    &= 2 \pi c_k \sum_{k' \in \Z} c_{k'} [z]^k[z']^{k'}.
\end{align*}

Then, replacing $z$ and $z'$ with wavelet coefficients, we get

\[
    \sum_{\al \in \mathcal{A}_A} e^{-ik\al} C^\ALPHA x((j,\theta,\al), (j',\theta',0),\tau) \simeq 2 \pi c_k \sum_{k'\in \Z}c_{k'} C^\WPH x((j,\theta,k),(j',\theta',k'),\tau).
\]

Using Plancherel's theorem, we can write that

\begin{align*}
    C^\WPH x((j,\theta,k),(j',\theta',k'),\tau) &= \frac{1}{|\omN|} \sum_{\om \in \frac{2\pi}{N}\omN} \mathcal{F}([x \star \psi_\la]^k)(\om) \mathcal{F}([x \star t_\tau\psi_\la]^{-k'})(\om),
\end{align*}
where the Fourier transform of an image $x$ is defined by $\mathcal{F}(x)(\om) := \sum_{u \in \omN} x(u) e^{-i\om u}$, and $t_\tau$ denotes the translation by $\tau$, i.e. $t_\tau f(\cdot) = f(\cdot - \tau)$.

Now, suppose that the wavelets $\psi_\al$ have disjoint compact frequency support, in balls $B_{\la}(2^{-j}C')$, where $\la=2^{-j}r_{-\theta}\wxi$, and $\wxi$ is the central frequency of the mother wavelet $\psi$ (cf. ?).
Suppose also that frequency transposition property of the phase harmonics operator (cf. \cite{mallat2020phase}) is such that $[x \star \psi_\la]^k$ has (approximately) frequency support in $B_{k\la}(k2^{-j}C')$. Then, for all $\la, \la'$, and all $k \in \Z$, there exists only one $k^\ast$ such that the frequency supports of $[x \star \psi_\la]^k$ and $[x \star \psi_{\la'}]^{-k^\ast}$ are not disjoint, i.e. such that $C^\WPH x((j,\theta,k),(j',\theta',k'),\tau) \not= 0$. This tells us that 
\[
\sum_{\al \in \mathcal{A}_A} e^{-ik\al} C^\ALPHA x((j,\theta,\al), (j',\theta',0),\tau) \simeq c_k c_{k^\ast} C^\WPH x((j,\theta,k),(j',\theta',k^\ast),\tau).
\]

Thus, computing all correlations for $\al \in \mathcal A_4$, and $\al' = 0$ gives us (approximately) all the information contained in WPH coefficients for any pair $k, k'$.

This result lies on several approximations, and strong assumptions about the wavelets, which are not fully met in practice. For this reason, setting $\al' = 0$ instead of $\al' \in \mathcal A_4$ effectively reduced the amount of information captured by the statistics, and therefore increases the diversity of the model. However, as we observe in \Cref{ss.extremes}, there is not too much information lost, and the resulting model still captures most of the important geometric structures in texture images.

\section{Reduced color model}\label{app.reduced-c}

One can reduce the number of statistics in the color model by selection the spatial shift parameter $\tau$ to be non-zero only for correlations between the same color channels.
More precisely, it is defined by the following index set:
$\U := \{ (\ga, \ga', \tau) : ((j, \theta, \al), (j', \theta', \al'), \tau) \in \U^{\ALPHAI}, c= c' \in \{1, 2, 3\}\} \cup \{ (\ga, \ga', 0) : ((j, \theta, \al), (j', \theta', \al'), 0) \in \U^{\ALPHAI}, (c, c') \in \{1, 2, 3\}^2 \}$.
This gives a model of size $\sim113$k, with little degradation of the visual quality, as shown in \Cref{fig:color-reduced}.

{\setlength{\tabcolsep}{2pt}
\begin{figure}[!h]
    \centering
    \begin{tabular}{ccc}
       Obs & $\ALPHAC$ (320k) & Reduced (113k) \\
       \includegraphics[width=0.2\linewidth]{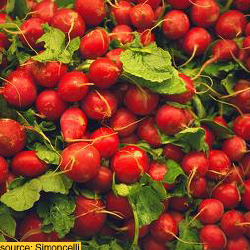}&
       \includegraphics[width=0.2\linewidth]{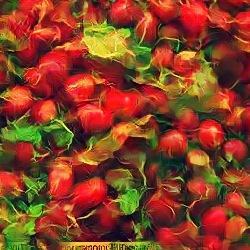}&
       \includegraphics[width=0.2\linewidth]{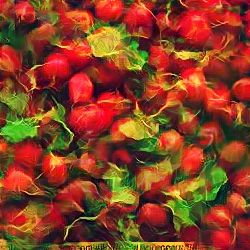}\\
       \includegraphics[width=0.2\linewidth]{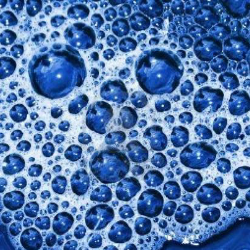}&
       \includegraphics[width=0.2\linewidth]{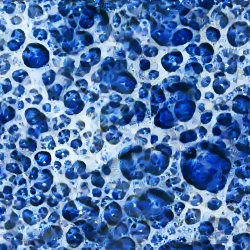}&
       \includegraphics[width=0.2\linewidth]{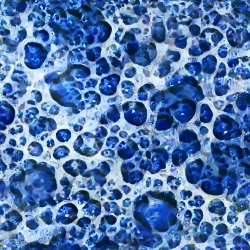}\\
       \includegraphics[width=0.2\linewidth]{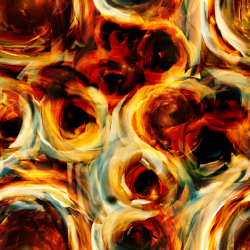}&
       \includegraphics[width=0.2\linewidth]{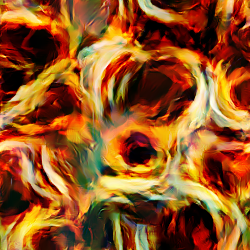}&
       \includegraphics[width=0.2\linewidth]{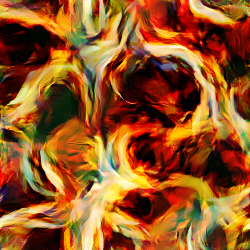}\\
      \includegraphics[width=0.2\linewidth]{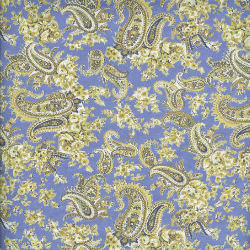}&
      \includegraphics[width=0.2\linewidth]{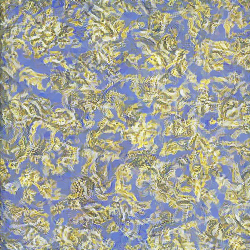}&
      \includegraphics[width=0.2\linewidth]{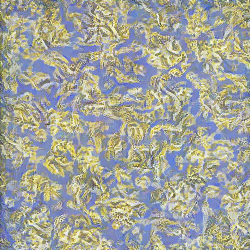}\\
      \includegraphics[width=0.2\linewidth]{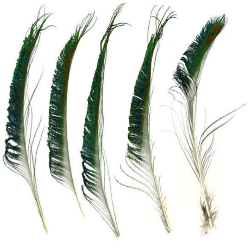}&
      \includegraphics[width=0.2\linewidth]{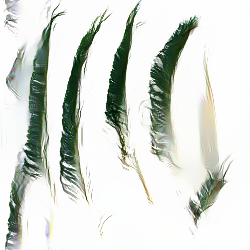}&
      \includegraphics[width=0.2\linewidth]{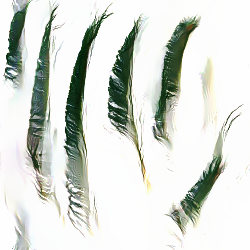}\\
       \includegraphics[width=0.2\linewidth]{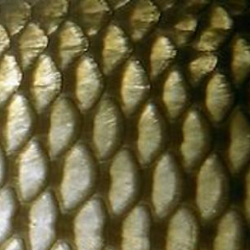}&
      \includegraphics[width=0.2\linewidth]{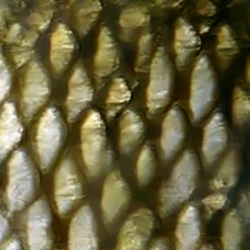}&
      \includegraphics[width=0.2\linewidth]{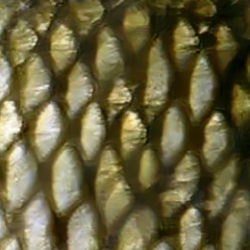}\\
    \end{tabular}
    \caption{Visual comparison of syntheses from $\ALPHAC$ model and its reduced version.}
    \label{fig:color-reduced}
\end{figure}
}

\section{VGG score}\label{app:vgg}

In \citet{ustyuzhaninov2017what}, the authors proposed to use the synthesis loss of the VGG model to evaluate the quality of syntheses from any model. The goal is to define a quantitative, and more objective evaluation method than mere visual inspection. Since the VGG model produces syntheses almost indistinguishable from real textures, it is natural to consider its loss to asses the quality of a synthesis.
We computed this loss for the first two examples of \Cref{fig:comp-ps-vgg} (radishes and cherries), and the frist two examples of \Cref{fig:gray} (gravel and Turbulence flow).
Note however that this loss is not exactly the same as the one used in \citet{ustyuzhaninov2017what}, as the layers selected to compute the loss are different. In this work, we chose to use the layers suggested in \citet{gatys2015texture}, (i.e. 'conv1\_1', 'pool1', 'pool2', 'pool3', and 'pool4' of the VGG-19 network \citep{simonyan2014very}), and compute the relative VGG loss\footnote{Using the code from \url{https://github.com/ivust/random-texture-synthesis/blob/master/vgg_loss.py} (function style\_loss\_relative).}. 

We notice that this score is not always consistent with visual inspection, as there are texture examples and models for which the syntheses do not look much like the observation image, yet produce a small VGG loss (see e.g. the first and last rows of \Cref{fig:comp-ps-vgg}, the RF model syntheses have the smallest loss). It should also be noted that the VGG loss reported on the VGG syntheses is {\em not} the synthesis loss after optimization, as a histogram matching (HM) procedure is performed as post-processing after optimization. We observed that the VGG loss of the syntheses from the VGG model after HM was considerably higher than the one for syntheses before it, while being visually very similar as illustrated in \Cref{fig:HM}.
These observations suggest that the VGG score suffers from instabilities after reaching a certain level (that is, if the VGG loss is small enough, small perturbations of the values of the image pixels might have a strong impact on the loss).

\begin{table}[!h]
    \centering
    \caption{Relative VGG loss of gray-scale textures for the first two examples of \Cref{fig:comp-ps-vgg}, and the first two examples of \Cref{fig:gray}.}
    \begin{tabular}{c|c|c|c|c}
      Data / Model & $\ALPHAI$  & VGG  & PS   & RF \\
        \hline
       Radishes    & 5.02e-05      &  1.87e-05  &   2.37e-04   &  1.13e-05 \\
        \hline
       Cherries   &  4.86e-05    &  1.47e-06  &  6.68e-04 &  9.65e-06 \\
       \hline
       Gravel     &  5.97e-05       &  3.08e-06  &  7.25e-04 & 1.29e-05 \\
        \hline
       Turbulence   & 5.59e-05   &  5.97e-05  & 2.42e-04 & 3.95e-05 \\ 
    \end{tabular}
    
    \label{tab:vgg-loss}
\end{table}

{\setlength{\tabcolsep}{2pt}
\begin{figure}[!h]
    \centering
    \begin{tabular}{cc}
       Before HM & After HM\\
       \includegraphics[width=0.2\linewidth]{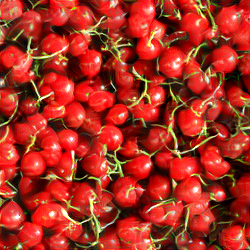}&
       \includegraphics[width=0.2\linewidth]{figs_color/vgg/cerise512_vgg.pdf}
    \end{tabular}
    \caption{Visual comparison of syntheses from the VGG model, before and after histogram matching. Before HM, the relative VGG loss is 2.22e-08, while after HM, the loss is 5.38e-05.}
    \label{fig:HM}
\end{figure}
}

\section{Influence of the choice of the wavelet transform}\label{app:wavelet}

\subsection{Influence of the wavelet family}
In Section \ref{ss.extremes}, we illustrated the importance of the set of indices $\U$ that define the wavelet coefficients being correlated. Another important role is played by the choice of the wavelets used in \eqref{periodic}.
As illustrated in \Cref{fig:wavelet-families}, this choice can have a visible impact on the quality of the textures.
We observe that, while on the first example, the coherence of the structures appear similar for the three wavelet families, the second example shows that the wavelets used in \citet{portilla2000parametric} are less efficient in reproducing the contours of the objects (pebbles). While in our experiments, we chose to use the classical Morlet wavelets, an optimal choice for the wavelet family remains an open problem.

{\setlength{\tabcolsep}{2pt}
\begin{figure}[!h]
    \centering
    \begin{tabular}{cccc}
         Observation & Simoncelli & Bump & Morlet \\
         \includegraphics[width=0.17\linewidth]{figs/obs/cerise_original2.pdf} &
         \includegraphics[width=0.17\linewidth]{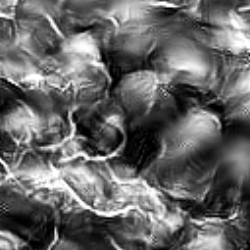}&
         \includegraphics[width=0.17\linewidth]{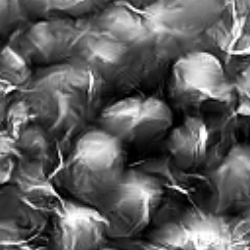}&
         \includegraphics[width=0.17\linewidth]{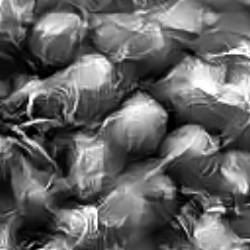}\\
         \includegraphics[width=0.17\linewidth]{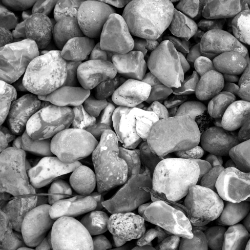}&
         \includegraphics[width=0.17\linewidth]{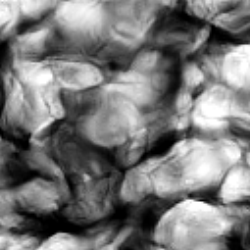}&
         \includegraphics[width=0.17\linewidth]{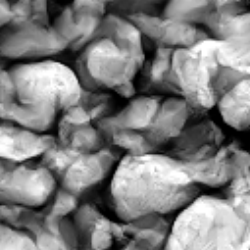}&
         \includegraphics[width=0.17\linewidth]{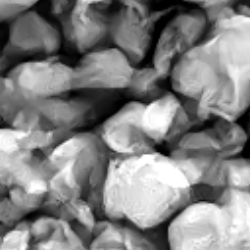}\\
    \end{tabular}
    \caption{
    Comparison between different wavelets families used in the $\ALPHAI$ model.
    Central zooms of syntheses using the same covariance model, with three different wavelet families. From left to right: observation, Simoncelli steerable wavelets, bump steerable wavelets \citep{mallat2020phase}, and Morlet wavelets.
    }
    \label{fig:wavelet-families}
\end{figure}}

\subsection{Influence of scale parameter}
\label{app:scale}

Recall from \Cref{ss.wavelet-transform}, the wavelet transform of an image $x$ is defined by 
\[
    \{ x \star \psi_{j, \theta}, \: x \star \phi_J\}_{0 \leq j < J, \: \theta \in \frac{\pi}{L}\{0, \cdots, L-1\}}.
\] 

The maximal scale parameter $J$ also plays an important role in the definition of the wavelet transform. It determines the scales of the structures being captured by the transform. If this parameter is too small, large structures in the observation image might not be captured and reproduced in the model syntheses. Conversely, if $J$ is too large, then the large scale statistics may have a high variance, inducing a memorization effect in the syntheses. \Cref{fig:scale} illustrates this point on two examples from \Cref{ss.results}. By setting $J=4$ (i.e. the maximal range of structures captured by the wavelets is of size $2^4=16$), we observe on the first example that the larger structures (bubbles) are not well reproduced. When $J$ is set to 6, the observation is almost identically reproduced by the synthesis. Similarly on the second examples, several parts of the synthesis appear very similar to ones in the observation. We found that a suitable trade-off consists in setting $J = 5$ for images of size $N=256$.

{\setlength{\tabcolsep}{2pt}
\begin{figure}[!h]
    \centering
    \begin{tabular}{cccc}
        Observation & $J=4$ & $J=5$ & $J=6$ \\
        \includegraphics[width=0.17\linewidth]{figs_color/obs/bubbly.pdf} &
        \includegraphics[width=0.17\linewidth]{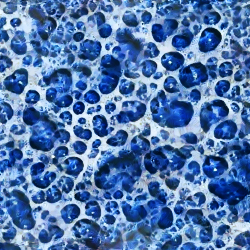} &
        \includegraphics[width=0.17\linewidth]{figs_color/alpha/bubbly_alpha_full_corr_morlet_L4_dj4_Aa41_hist.pdf}&
        \includegraphics[width=0.17\linewidth]{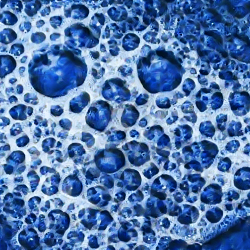}\\
        \includegraphics[width=0.17\linewidth]{figs_color/obs/cerise.pdf} &
        \includegraphics[width=0.17\linewidth]{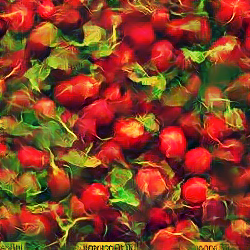} &
        \includegraphics[width=0.17\linewidth]{figs_color/alpha/cerise_alpha_full_corr_morlet_L4_dj4_Aa41_hist.pdf}&
        \includegraphics[width=0.17\linewidth]{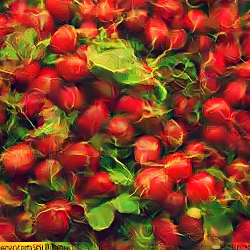}\\
    \end{tabular}
    \caption{Syntheses from the $\ALPHAC$ model defined with three different maximal scale parameters $J \in \{4,5,6\}$. As in \Cref{s.results}, Morlet wavelets are used.
    }
    \label{fig:scale}
\end{figure}
}

\section{Supplementary results of $\ALPHAI$/$\ALPHAC$ vs. PS, RF, VGG}\label{supp:comp}

In \Cref{fig:gray} we present additional syntheses on various examples, from the PS, $\ALPHAC$, RF and VGG models. These examples can be viewed as random (Turbulence flow, tree bark, porous stone), structured (gravel, paisley pattern, tree leaf, school text), or inhomogeneous (crafted pattern of third row, but also the porous stone). 

We see that $\ALPHA_I$ again significantly 
improves the visual quality of the PS model on gray-scale textures such as the gravel and turbulence flow. 
The visual quality of RF and VGG seems also worse on Turbulence flow compared to $\ALPHA_I$. 
In some examples such as tree leaf, we find the synthesis of all the models are similar. 
On the inhomogeneous porous stone, non of the models give satisfying visual results. 


Similarly, in \Cref{fig:color} are presented supplementary syntheses form the color models, for structured images (radishes, bubbles, flowers), quasi-periodic images (scales, honeycomb, bricks), and non-stationary images (feathers). As previously observed, for highly structured quasi-periodic images such as the bricks example, the VGG model fails to capture long-range correlation, which can be solved using the method of \citet{berger17incorp}. Syntheses of non-stationary images exhibit memorization effects, as previously observed.

{\setlength{\tabcolsep}{2pt}
\begin{figure}[!h]
    \centering
    \begin{tabular}{ccccc}
        Observation & PS (3.2k) & RF (525k) & $\ALPHAI$ (35k) & VGG (177k)\\
         \includegraphics[width=0.17\linewidth]{figs/obs/turb_zoom.pdf}&
        \includegraphics[width=0.17\linewidth]{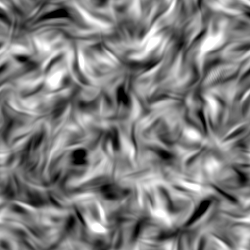}&
        \includegraphics[width=0.17\linewidth]{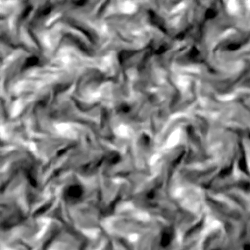}&
        \includegraphics[width=0.17\linewidth]{./figs/alpha/turb_zoom_alpha_all_morlet_L4_dj4_Aa41_hist.pdf}&
         \includegraphics[width=0.17\linewidth]{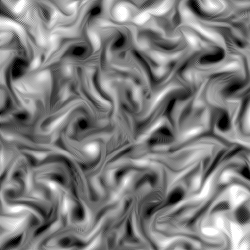}\\
         \includegraphics[width=0.17\linewidth]{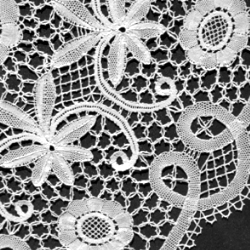}&
        \includegraphics[width=0.17\linewidth]{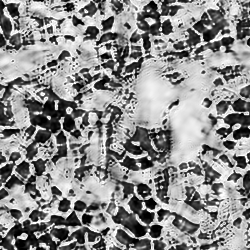}&
        \includegraphics[width=0.17\linewidth]{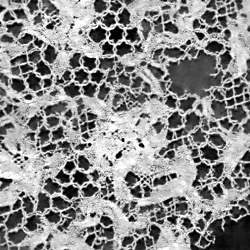}&
        \includegraphics[width=0.17\linewidth]{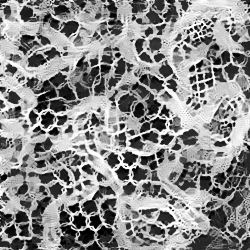}&
         \includegraphics[width=0.17\linewidth]{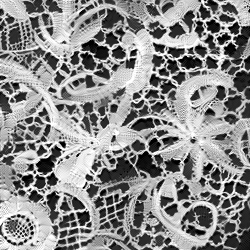}\\
         \includegraphics[width=0.17\linewidth]{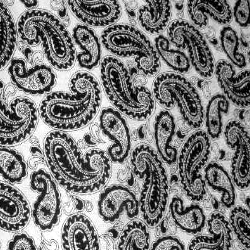}&
        \includegraphics[width=0.17\linewidth]{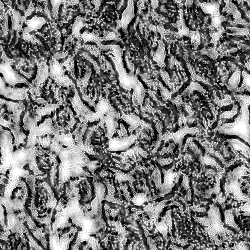}&
        \includegraphics[width=0.17\linewidth]{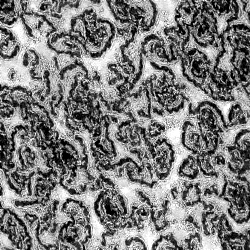}&
        \includegraphics[width=0.17\linewidth]{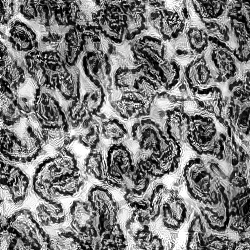}&
         \includegraphics[width=0.17\linewidth]{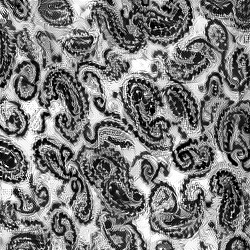}\\
         \includegraphics[width=0.17\linewidth]{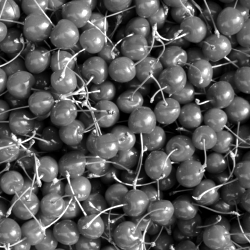}&
        \includegraphics[width=0.17\linewidth]{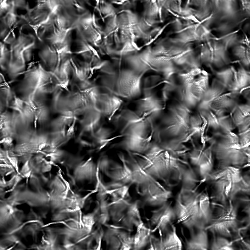}&
        \includegraphics[width=0.17\linewidth]{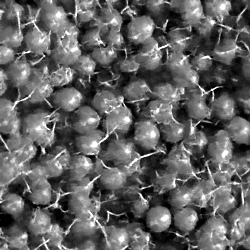}&
        \includegraphics[width=0.17\linewidth]{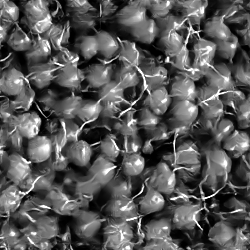}&
        \includegraphics[width=0.17\linewidth]{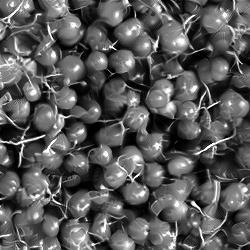}\\
         \includegraphics[width=0.17\linewidth]{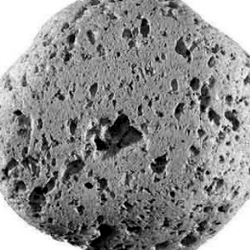}&
        \includegraphics[width=0.17\linewidth]{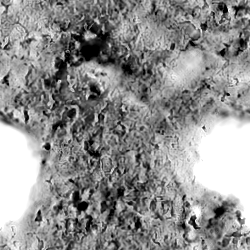}&
        \includegraphics[width=0.17\linewidth]{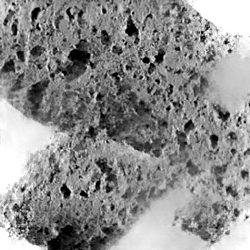}&
        \includegraphics[width=0.17\linewidth]{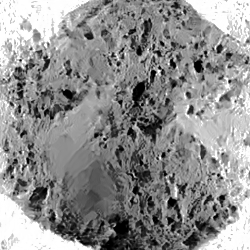}&
         \includegraphics[width=0.17\linewidth]{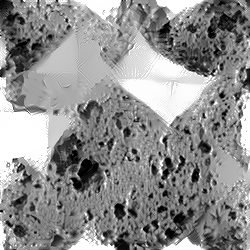}\\
         \includegraphics[width=0.17\linewidth]{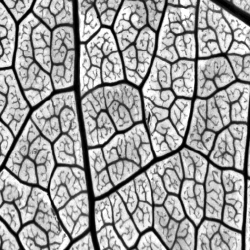}&
        \includegraphics[width=0.17\linewidth]{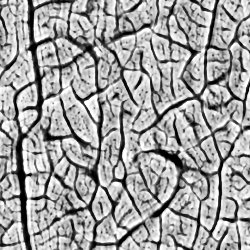}&
        \includegraphics[width=0.17\linewidth]{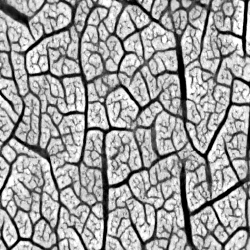}&
        \includegraphics[width=0.17\linewidth]{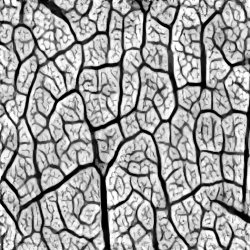}&
         \includegraphics[width=0.17\linewidth]{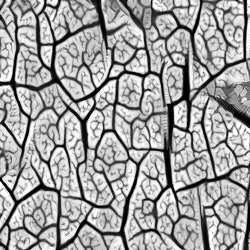}\\
         \includegraphics[width=0.17\linewidth]{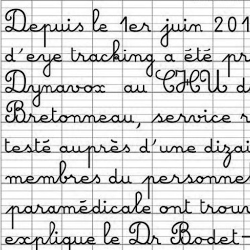}&
        \includegraphics[width=0.17\linewidth]{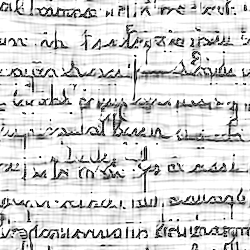}&
        \includegraphics[width=0.17\linewidth]{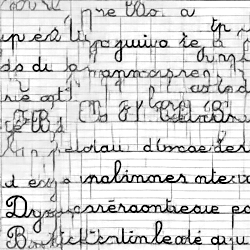}&
        \includegraphics[width=0.17\linewidth]{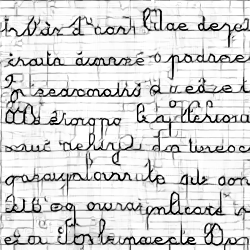}&
         \includegraphics[width=0.17\linewidth]{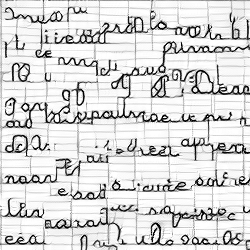}\\
         \includegraphics[width=0.17\linewidth]{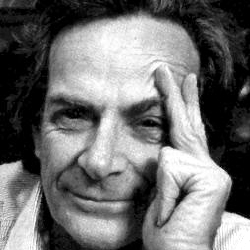}&
        \includegraphics[width=0.17\linewidth]{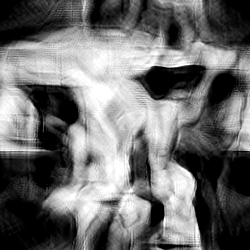}&
        \includegraphics[width=0.17\linewidth]{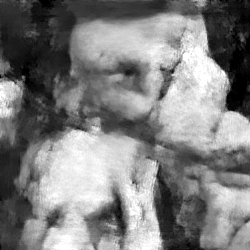}&
        \includegraphics[width=0.17\linewidth]{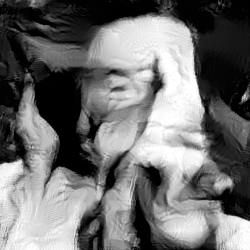}&
        \includegraphics[width=0.17\linewidth]{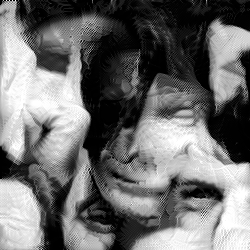}\\
    \end{tabular}
    \caption{Visual comparison between different texture models on gray-scale images.}
    \label{fig:gray}
\end{figure}}

{\setlength{\tabcolsep}{2pt}
\begin{figure}[!h]
    \centering
    \begin{tabular}{ccccc}
    Observation & PS (17k) & RF (525k) & $\textstyle{\ALPHAC}$(320k) & VGG (177k)\\
        \includegraphics[width=0.17\linewidth]{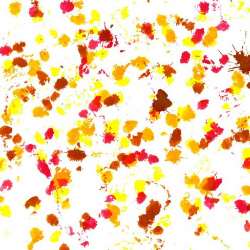}&
        \includegraphics[width=0.17\linewidth]{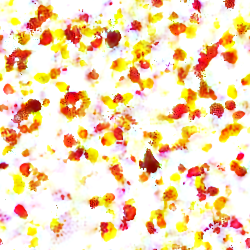}&
        \includegraphics[width=0.17\linewidth]{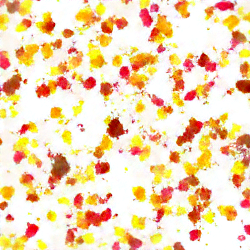}&
        \includegraphics[width=0.17\linewidth]{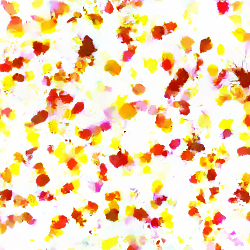}&
        \includegraphics[width=0.17\linewidth]{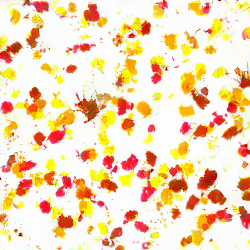}\\
        \includegraphics[width=0.17\linewidth]{figs_color/obs/cerise.pdf}&
        \includegraphics[width=0.17\linewidth]{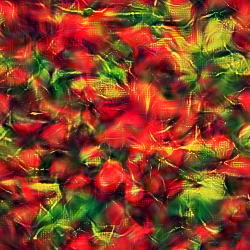}&
        \includegraphics[width=0.17\linewidth]{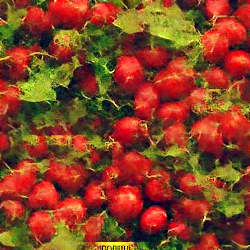}&
        \includegraphics[width=0.17\linewidth]{figs_color/alpha/cerise_alpha_full_corr_morlet_L4_dj4_Aa41_hist.pdf}&
        \includegraphics[width=0.17\linewidth]{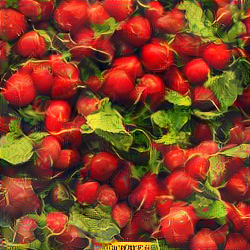}\\
        \includegraphics[width=0.17\linewidth]{figs_color/obs/bubbly.pdf}&
         \includegraphics[width=0.17\linewidth]{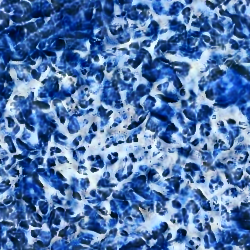}&
         \includegraphics[width=0.17\linewidth]{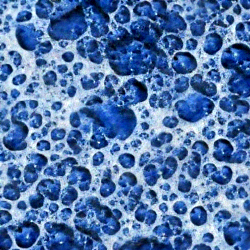}&
         \includegraphics[width=0.17\linewidth]{figs_color/alpha/bubbly_alpha_full_corr_morlet_L4_dj4_Aa41_hist.pdf}&
         \includegraphics[width=0.17\linewidth]{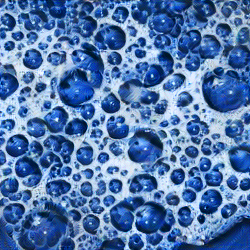}\\
         \includegraphics[width=0.17\linewidth]{figs_color/obs/scaly_scaly.pdf}&
         \includegraphics[width=0.17\linewidth]{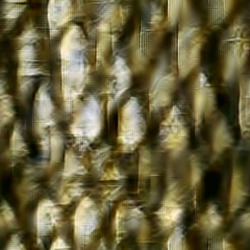}&
         \includegraphics[width=0.17\linewidth]{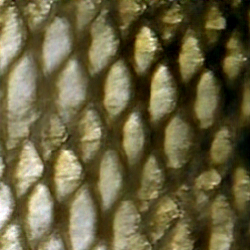}&
         \includegraphics[width=0.17\linewidth]{figs_color/alpha/scaly_alpha_full_corr_morlet_L4_dj4_Aa41_hist.pdf}&
         \includegraphics[width=0.17\linewidth]{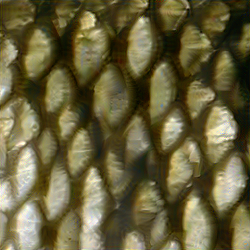}\\
         \includegraphics[width=0.17\linewidth]{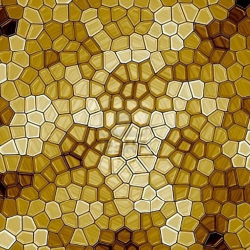}&
         \includegraphics[width=0.17\linewidth]{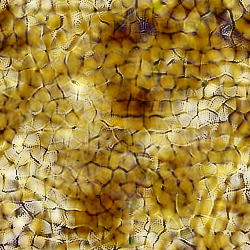}&
         \includegraphics[width=0.17\linewidth]{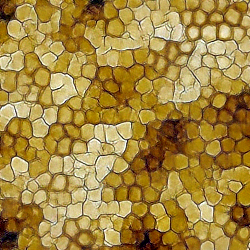}&
         \includegraphics[width=0.17\linewidth]{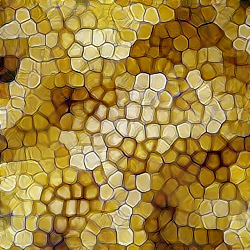}&
         \includegraphics[width=0.17\linewidth]{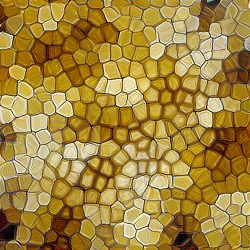}\\
         \includegraphics[width=0.17\linewidth]{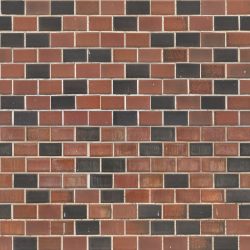}&
        \includegraphics[width=0.17\linewidth]{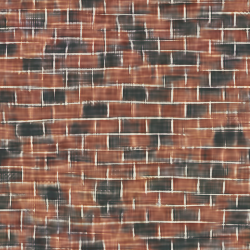}&
        \includegraphics[width=0.17\linewidth]{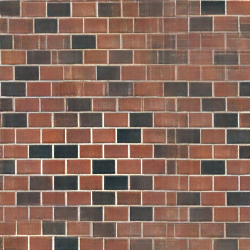}&
        \includegraphics[width=0.17\linewidth]{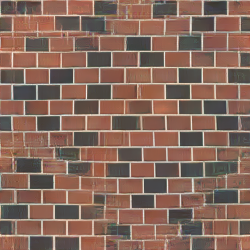}&
        \includegraphics[width=0.17\linewidth]{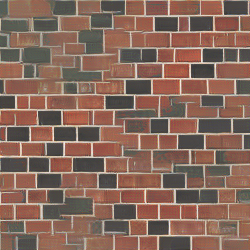}\\
         \includegraphics[width=0.17\linewidth]{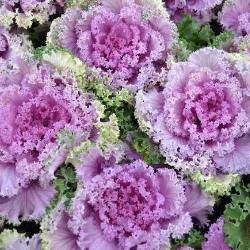}&
         \includegraphics[width=0.17\linewidth]{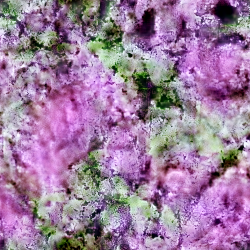}&
         \includegraphics[width=0.17\linewidth]{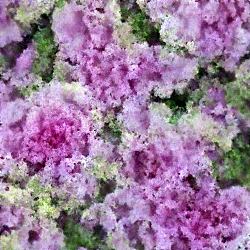}&
         \includegraphics[width=0.17\linewidth]{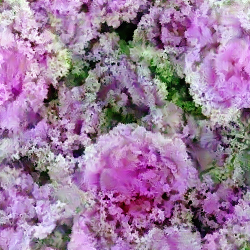}&
         \includegraphics[width=0.17\linewidth]{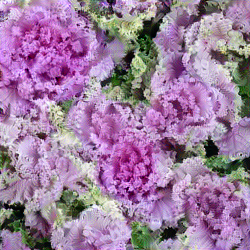}\\
         \includegraphics[width=0.17\linewidth]{figs_color/obs/feathers_feathers.pdf}&
         \includegraphics[width=0.17\linewidth]{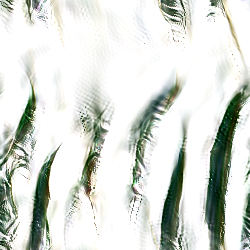}&
         \includegraphics[width=0.17\linewidth]{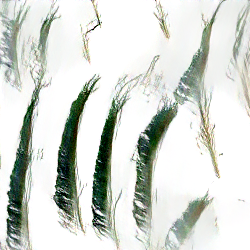}&
         \includegraphics[width=0.17\linewidth]{figs_color/alpha/feathers_alpha_full_corr_morlet_L4_dj4_Aa41_hist.pdf}&
         \includegraphics[width=0.17\linewidth]{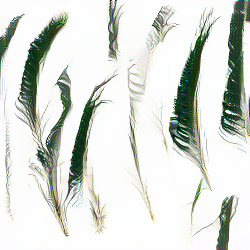}\\

    \end{tabular}
    \caption{Visual comparison between different texture models on color images.}
    \label{fig:color}
\end{figure}}

\end{document}